\documentclass[12pt]{article}
\usepackage{amsmath}
\usepackage{graphicx}
\usepackage{color}
\usepackage{multirow}
\usepackage[natbibapa]{apacite}
\usepackage{rotating}
\usepackage{bbm}
\usepackage{latexsym}

\usepackage{times}
\usepackage{epsfig}
\usepackage{graphicx}
\usepackage{amsmath}
\usepackage{amssymb}
\usepackage{soul}
\usepackage{capt-of}
\usepackage{multirow}
\usepackage{adjustbox}
\usepackage[thinlines]{easytable}
\usepackage{gensymb}
\usepackage{booktabs}
\usepackage[justification=raggedright]{caption} % TODO: double check this
\usepackage{subcaption}
\usepackage{mfirstuc}

\usepackage[pagebackref=true,breaklinks=true,colorlinks,bookmarks=false,letterpaper=true]{hyperref}
\newcommand{\todo}[1]{}
\newcommand{\IA}[1]{intelligent augmentation}
\newcommand{\AD}[1]{adversarial independence}
\newcommand{\HI}[1]{TARA}
\newcommand{\BT}[1]{Best Thresholding}

\newcommand{\captionfonts}{\normalsize}

\makeatletter  
\long\def\@makecaption#1#2{%
  \vskip\abovecaptionskip
  \sbox\@tempboxa{{\captionfonts #1: #2}}%
  \ifdim \wd\@tempboxa >\hsize
    {\captionfonts #1: #2\par}
  \else
    \hbox to\hsize{\hfil\box\@tempboxa\hfil}%
  \fi
  \vskip\belowcaptionskip}
\makeatother   
\begin{document}
\hspace{13.9cm}%1

\ \vspace{20mm}\\

{\LARGE \centering TARA: Training and Representation Alteration \\ for AI Fairness and Domain Generalization}

{\bf \large William Paul$^{\displaystyle 1}$, Armin Hadzic$^{\displaystyle 1}$, Neil Joshi $^{\displaystyle 1}$, Fady Alajaji$^{\displaystyle 2}$, Philippe Burlina$^{\displaystyle 1, 3}$}\\
{$^{\displaystyle 1}$The Johns Hopkins University Applied Physics Laboratory\\
11100 Johns Hopkins Rd,
Laurel, MD 20723, USA\\
{\tt\small {firstname.lastname}@jhuapl.edu}}\\
{$^{\displaystyle 2}$Department of Mathematics and Statistics\\ Queens University, ON K7L 3N6, Canada \\
{\tt\small {fa}@queensu.ca}}\\
{$^{\displaystyle 3}$Department of Computer Science\\ Johns Hopkins University\\
3400 N. Charles Street
Baltimore, MD 21218\\}

%\ \\[-2mm]

\thispagestyle{empty}
\markboth{}{NC instructions}
\ \vspace{-0mm}\\
%
%Abstract
\begin{center} {\bf Abstract} \end{center}
%limit 400 words!!
%While deep learning (DL) approaches are reaching human-level performance for many tasks, attention is now shifting towards challenges affecting DL deployment, 
We propose a novel method for enforcing AI fairness with respect to protected or sensitive factors. 
%(e.g. sex, ethnicity, age). 
This method uses a dual strategy performing training and representation alteration (TARA) for the mitigation of prominent causes of AI bias by including: a) the use of representation learning alteration via \AD{} to suppress the bias-inducing dependence of the data representation from protected factors; and b) training set alteration via \IA{} to address bias-causing data imbalance, by using generative models that allow the fine control of sensitive factors related to underrepresented populations via domain adaptation and latent space manipulation. When testing our methods on image analytics, experiments demonstrate that \HI{} significantly or fully debiases baseline models while outperforming competing debiasing methods that have the same amount of information, e.g., with (\% overall accuracy, \% accuracy gap) = (78.8, 0.5) vs. the baseline method's score of  (71.8, 10.5) for EyePACS, and 
%trained on a dataset biased with regards to gender, 
(73.7, 11.8) vs. (69.1, 21.7) for CelebA. Furthermore, recognizing certain limitations in current metrics used for assessing debiasing performance, we propose novel conjunctive debiasing metrics. Our experiments also demonstrate the ability of these novel metrics in assessing the Pareto efficiency of the proposed methods.
%PB: Experiments show that those methods in some cases achieve debiasing even in the case of protected factor generalization.
%%%%%%%%%%%

%%%%%%%%% BODY TEXT
\section{Introduction}
Recent advances in AI, via deep learning, for tasks such as object detection (\cite{redmon2018yolov3}), retinal semantic segmentation (\cite{pekala2019deep}), or skin diagnostics (\cite{burlina2019automated}) have led to performance exceeding that of classical machine learning (\cite{burlina2011automatic}), even reaching human-level performance. However, this success is tempered by challenges such as private information leakage (\cite{shokri2017membership}), adversarial attacks (\cite{carlini2017adversarial}), low shot learning (\cite{ravi2016optimization,burlina2020low}), or bias with regard to sensitive factors and protected subpopulations (\cite{burlina2020addressing}). These challenges threaten to derail AI deployment in many areas including healthcare, autonomy, or smart cities. In this work, we focus on addressing AI bias. 

Two of the dominant sources of AI bias include: (a) data disparity or imbalance with respect to protected subpopulation(s) and (b) conditional dependence of model predictions on protected factor(s). We report on novel approaches for addressing these sources of bias, tackling source (a) via generative methods that synthesize more data for underrepresented populations, while allowing for control of specific semantic attributes of images (called \IA{} or IA), in an approach that is also related to domain adaptation. We address source (b) via adversarial two player models that aim to minimize conditional dependence of the model prediction on protected factors (called \AD{} or AD). Since models may be affected by both sources of bias, we investigate a novel method jointly exploiting the above two strategies consisting of {\it training and representation alteration} (termed \HI{}). Finally, the problem of fairness is also  related to generalization~\cite{hu2018does}  and~\cite{sagawa2019distributionally}. In our study, rather than typical settings that use the natural proportions in the dataset, we focus on an extreme case of data bias akin to domain generalization, where the minority subpopulation, defined as the combination of protected and target classes with the fewest data points, is entirely excluded.
%\todo{(Lux. later) Replace Table 1 by splash describing the contents of Table 1, and fix corresponding sentence referencing Table 1. Move Table 1 to appendix as well.}

\section{Prior Work} 
%State of the art (SOTA) prior work include:
{\em A Taxonomy of Methods:}  A number of recent studies have investigated AI fairness, for instance~\cite{zemel2013learning} studies demographic parity,~\cite{bolukbasi2016man} tackles debiasing of word embeddings,~\cite{prost2019toward} and~\cite{zemel2013learning} explore distribution matching to maximize fairness, and~\cite{kinyanjui2020fairness} presents a method for medical image debiasing for skin segmentation. Generally speaking, there are three main intervention methods to mitigate bias and promote fairness in deep learning~\cite{caton2020fairness}: (1) pre-processing techniques, including among others, sampling, variable-masking and generative models, which attempt to augment, repair or balance training data that is biased vis-a-vis sensitive or protected factors; (2) in-processing techniques, such as adversarial learning, regularization and constrained optimization, which 
actively alter the model at hand by introducing fairness metrics in the optimization and finding a right balance between competing objectives (e.g., performance vs fairness); (3) post-processing techniques, such as calibration and thresholding, which aim to ensure fair outcomes (e.g., prediction outcomes) upon realizing that the implemented method can result in outputs that are biased to some sensitive  attributes. The reader is referred to~\cite{caton2020fairness} for a detailed inventory of prior papers on the above approaches. We focus on two of these fairness methods, one based on (in-processing) adversarial learning, and the other based on (pre-processing) generative augmentation, which we discuss in further detail in the following paragraphs.

{\em In-Processing Adversarial Approaches:} Several studies addressing the {\em conditional dependence} source of bias employ adversarial methods (\cite{goodfellow2014generative}). Enforcing fairness in domain adaption in~\cite{ganin2016domain} uses this strategy. 
\cite{beutel2017data} uses a separate adversarial network in a natural language processing task to predict the protected factor, and modifies the word embeddings to reduce the adversary's performance. Similarly,~\cite{alvi2018turning} employs multiple network heads and a cross entropy loss comparing the predicted distribution to a uniform distribution to reduce bias in embeddings across multiple protected factors in the image domain. The concurrent studies \cite{wadsworth2018achieving} and \cite{zhang2018mitigating}, also partly an inspiration for our adversarial method, use a similar approach and expand to tabular data targeting various forms of fairness.
Also,~\cite{song2019learning} uses an information theoretic approach to mitigate bias through fair controllable representations of data and sensitive factors on tabular data. 
The method from~\cite{wang2019balanced} shows that even balanced datasets can exhibit bias and that using adversarial methods to mask out markers of protected factors (gender) directly in the image domain can provide benefits in some cases. ~\cite{edwards2015censoring}'s approach uses an adversarial independence approach to maximize fairness of an autoencoder internal representation, which is somewhat related to our own method, but differs from it in that it maximizes the less strict demographic parity instead of equality of odds.
These findings motivate our approach. However, we depart here from these studies in several important ways: our adversarial network feeds off the internal representation to further reduce protected factor information leakage to the adversarial network, and we combine this adversarial approach with a novel augmentation approach that allows for selective image marker alteration. Additionally, we consider more extreme cases of bias where the minority population is not represented at all in the dataset.

{\em Pre-Processing, Augmentation and Generative Methods:}  Among methods that implement pre-processing and training data alteration, we note the approach by~\cite{quadrianto2019discovering} which uses a method of bias reduction by performing an image translation, exploiting a highly unconstrained mapping to latent space via residual statistics and enforcing equality of outcome in visual features of faces. \cite{hwang2020fairfacegan} tries to make image translation preserve protected attributes through utilizing a variational autoencoder as a generator in GANs, and evaluates the effectiveness of this translation via data augmentation like this work. Also related to our work, \cite{sattigeri2018fairness} uses GANs to generate debiased datasets, but does so by encouraging fairness of the generated data through the discriminator during training rather than optimizing the latent codes as we will demonstrate here.

To address the other important cause of bias, i.e., data imbalance, our strategy for debiasing also leverages \IA{}, which uses generative models that generate more data but allow fine control of image attributes for underrepresented factors. 
Generative approaches applicable to generating more synthetic data include:
generative models such as GANs~(\cite{karras2019style,grover2019bias}), autoencoders(~\cite{madras2018learning}), variational autoencoders (VAEs)~(\cite{kingma2013auto, louizos2015variational}) and generative autoregressive models, invertible flow-based latent vector models, or a hybrid of such models \cite{zhao2017infovae}. Such methods have limitations for addressing data imbalance and bias: while  they generate realistic images, they do not allow for controlling images with specific attributes (e.g., ~\cite{burlina2019assessment} or \cite{karras2019style}), 
which would correspond to an underrepresented population (e.g., images of dark skin individuals with Lyme disease,~\cite{burlina2020ai}). This motivates the need for methods that allow fine control of individual semantic  factors. 
Along with this control, there is also the matter of ensuring those other uncontrolled attributes remain invariant, which arguably requires {\it disentanglement}. Control and disentanglement are related (but distinct) concepts formally defined via information theoretic measures~(\cite{paul2020unsupervised}). \cite{paul2020unsupervised} shows that optimizing the control of semantic attributes also promotes (theoretically and empirically) {\it disentanglement} among latent factors. Disentanglement appears to have an incidence in promoting fairness as shown empirically in~\cite{Locatello2019OnTF}; however this work  does not provide a method to achieve this.

Likewise, generative methods such as~\cite{paul2020unsupervised} that allow semantic control are interesting for intelligent augmentation but are in practice, more challenging for debiasing, due to the fact that such methods hinge on the discovery of factors that are aligned with protected factors (e.g., sex) and also because of the residual entanglement that would remain in the controlling latent space codes~\cite{paul2020unsupervised}.

Much of the difficulty of ensuring model fairness has to do with resilience to distributional shift and domain shift, which call for approaches to domain adaptation. Many generative methods have been proposed to address domain adaptation. Some examples of these methods include CycleGAN~\cite{cyclegan2017} or StarGAN~\cite{stargan2018} and pix2pix~\cite{pix2pix2017, pix2pix2019}.

In sum, while much has been done in past generative modeling, the  aforementioned limitations have motivated the development of our novel debiasing approach, which, is able to perform fine semantic attribute control for intelligent augmentation by using latent space manipulation methods, and also importantly enables control of such attributes while keeping other factors of variations fixed, thereby addressing entanglement.

\section{Novel Contributions}
Our novel contributions are therefore as follows:
\begin{enumerate}
    \item {\em Approach:} 
We introduce a new debiasing strategy, that is able to perform intelligent augmentation by exploiting a novel latent space manipulation method which can finely control data attributes, and also adds a newly formulated adversarial two player approach for enforcing conditional independence  working off the pre-logits layer of the classification network. We call this overall approach~\HI{}. This method is able to address, for the first time, dual sources of bias (imbalance and dependency) via the combined alteration of training data and data representation. This method is shown to significantly outperform competing debiasing strategies that use the same amount of information.
\item  {\em Metrics:} We identify and address certain shortcomings of current fairness metrics by proposing novel metrics and demonstrating their utility.
\item  {\em Generalization:} We demonstrate the ability to debias in scenarios of extreme data imbalance entailing domain generalization, which to our knowledge, has not been widely addressed in the AI fairness literature, where models completely lacked training data for specific subpopulations (e.g., dark skin individuals with retinal diseases). This is particularly important for several reasons:  subpopulations with combinations of factors (e.g., race/age/gender)  yield partitions with little or no training data. Furthermore, it addresses domain transfer when a training dataset was curated for a certain population (e.g., a diabetic retinopathy detector developed for the US population) and is deployed to a new domain (e.g., Singapore) now including a new ethnicity (e.g., Malais) not contained in the initial population.
\item {\em Proxy Measure for Sensitive Attributes and its Robustness:} We start examining the robustness of a debiasing approach to mismatch in sensitive attributes. Specifically, we propose for the first time the use of a proxy factor (i.e., the Individual Typology angle or ITA) for a sensitive attribute and demonstrate transfer of debiasing faculty from a proxy protected factor to a target factor.

Finally, as a contribution to medical imaging, we demonstrate, for the first time, the use of ITA as a relevant protected factor for retinal images, allowing for debiasing without requiring costly and error-prone manual clinical image annotation. 
\end{enumerate}
\section{Methods}

\noindent {\bf  Nomenclature and Definitions of Fairness:} Henceforth as nomenclature we denote protected factor(s) by the random variable $S$, the classifier's prediction by $\hat{Y}$, and the underlying true label by $Y$. Developing unbiased AI systems requires a clear understanding of what constitutes fairness, which is ideally expressed in formal mathematical terms. Common definitions of fairness~(\cite{mehrabi2019survey,hardt2016}) include {\em demographic parity, equality of odds, and equality of opportunity} (see Appendix for mathematical formulations). All these formal definitions entail some form of conditional independence of the prediction $\hat{Y}$ from the sensitive protected factor $S$. Equality of odds, in particular, states that a predictive model must produce predictions that are conditionally independent of protected factors given the true outcome:
\begin{equation}
P(\hat{Y} = \hat{y} | S = s, Y = y)=P(\hat{Y} = \hat{y} | Y = y), \forall s, y, \hat{y}. 
\end{equation}

This motivates a method of debiasing that directly learns a data representation that exhibits conditional independence, an ingredient we use in the \AD{} and \HI{} methods. Our study adopts the stricter goal of equality of odds, which yields equality of performance (accuracy). This leads us to measure debiasing performance using  commonly adopted metrics such as accuracy and area under the receiver operating characteristic curve (AUC). We also propose novel metrics that are consistent with this goal, but address certain limitations of accuracy, which are detailed later.

We begin by introducing the \AD{} method for minimizing bias by maximizing conditional independence. Next, we present \IA{}, a method addressing data generation for underrepresented classes, which we then combine with \AD{} to form the \HI{} method. 

\todo{Can we change the classifier's variable name in Figure 1a from Y to F?}
\begin{figure}[tb!]
    \centering
    \begin{subfigure}[b]{0.48\textwidth}
        \centering
        %\vspace{-1.5cm}
        \includegraphics[width=\textwidth]{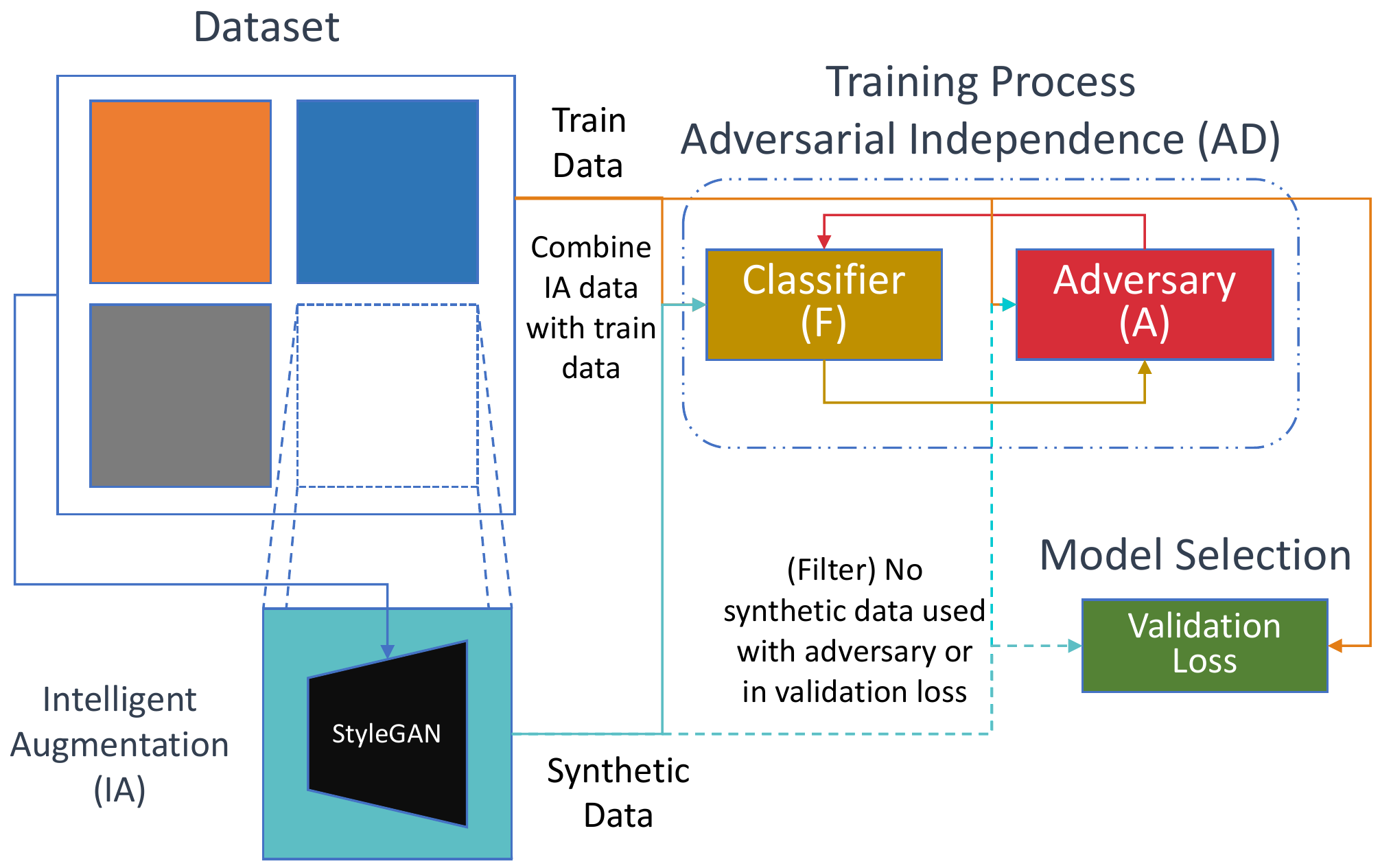}
        %\vspace{-3.5cm}
        \caption{TARA Process}
        
    \label{fig:TARA_network_diagram}
    \end{subfigure}
    \begin{subfigure}[b]{0.48\textwidth}
        \centering
        %\vspace{-1.5cm}
        \includegraphics[width=\textwidth]{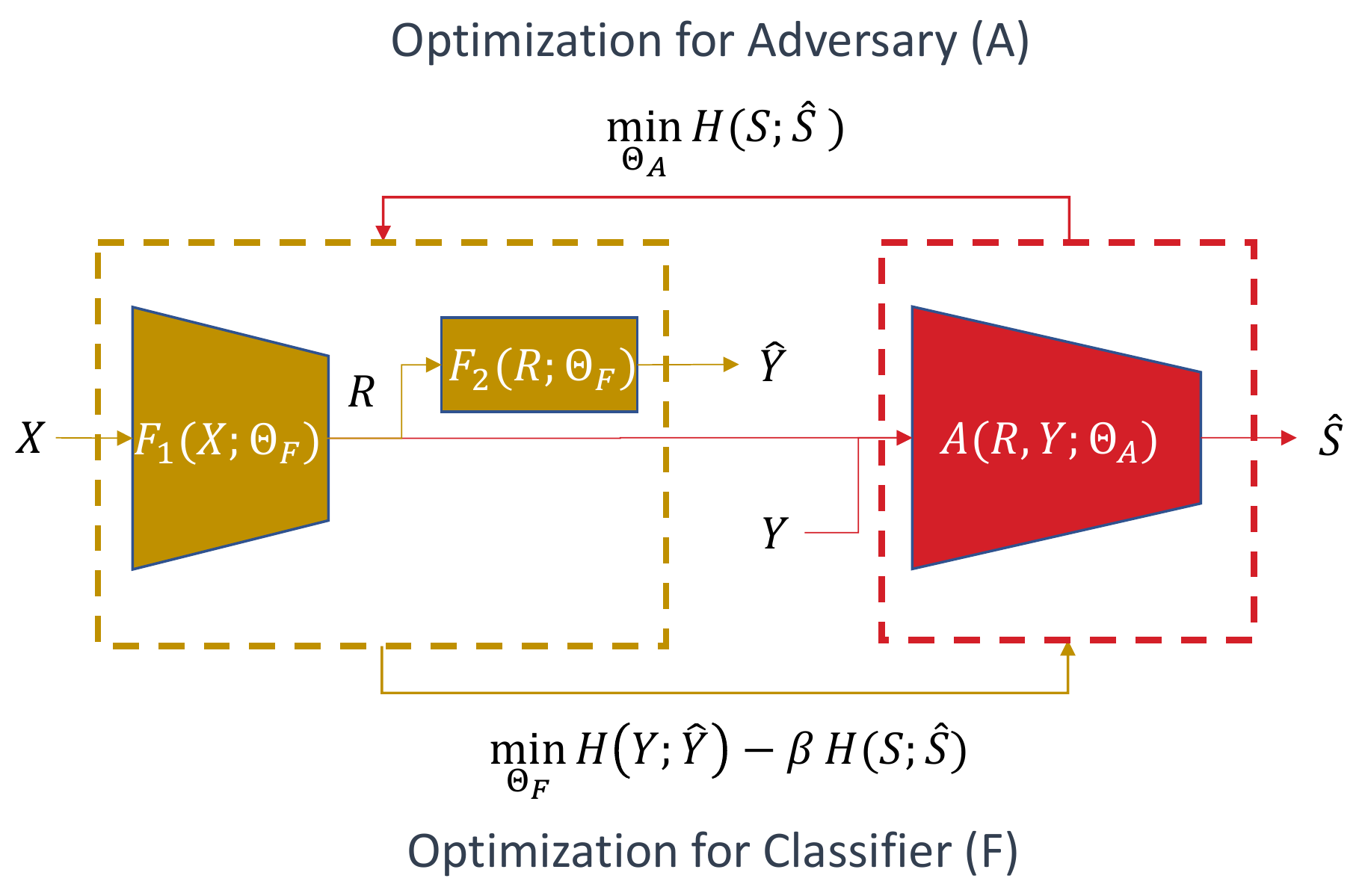}
        %\vspace{-3.5cm}
        \caption{Adversarial Independence}
        
    \label{fig:AD_network_diagram}
    \end{subfigure}
    \caption{ (a) The TARA process performs~\IA{} to generate samples for a subpopulation with no representation in the dataset. The resulting data can be used to train the adversarial module ($A$), described in (b), for~\AD{} training and classifier $F$ model selection. Represented by the teal dashed lines, the generated samples can be optionally filtered out of the input to the adversary in all phases of training and out of the model selection process (validation loss). (b) Bias is reduced by maximizing the prediction module's performance while minimizing the adversarial module's ($A$) ability to predict the protected factor ($S$) using the internal representation $R$ and known label ($Y$).}
\end{figure} %level one, and mild DR

\noindent {\bf Adversarial Independence:} A strategy for achieving resilience to bias is to learn a data representation where information about a protected factor can be suppressed. We use a method for debiasing neural network models using adversarial training derived from~\cite{zhang2018mitigating}. During training, we simultaneously train a prediction network $F$ with input $X$ and parameters $\Theta_F=(\Theta_{F_1},\Theta_{F_2})$ on a classification task represented via
$$F(X ; \Theta_F) = F_2(F_1(X; \Theta_{F_1}) ; \Theta_{F_2}) = F_2(R; \Theta_{F_2}) = \hat{Y}$$ and an adversarial network $A$ that aims to predict the protected factor $S$, as illustrated in Figure~\ref{fig:TARA_network_diagram}. In the process of computing $\hat{Y}$, the prediction network calculates some internal representation $R= F_1(X; \Theta_{F_1})$ that should be fair (chosen based on $F$ and $F_2$).
We consider two cases for the internal representation $R$:
\begin{itemize}
    \item For tabular data, we take $R$ corresponding to the logits with $F_2(R; \Theta_{F_2})=\textit{softmax}(R)$. 
    \item For image data, we take $R$ as the output of the flattening operation and prior to the final linear layer (for ResNet50, the architecture used in this paper), with $F_2(R; \Theta_{F_2})$ being a linear layer followed by a softmax layer. 
\end{itemize}
Simultaneously, the adversarial network with parameters $\Theta_A$ ingests this internal representation $R$ and the true $Y$ to produce $\hat{S}$ (the prediction for $S$):
$$A(R, Y; \Theta_A) = \hat{S}.$$ 
The cross entropy $H(\cdot;\cdot)$ is\footnote{Recall that, given two random variables $U$ and $V$ with distributions $P_U$ and $P_V$, respectively, then the cross-entropy between $U$ and $V$ is denoted by $H(U;V)$ and is given by $H(U;V)= \mathbb{E}_{P_U}\left[-\log P_V(U)\right]$, where $\mathbb{E}_{P_U}[\cdot]$ denotes expectation under the $P_U$ distribution.}
then applied to each of the two predictions and combined to compute the total loss which is optimized as follows:
\begin{equation}
    \min_{\Theta_F} \max_{\Theta_A} H(Y; F(X ; \Theta_F) ) - \beta H(S; A(R, Y; \Theta_A))
    \label{adv}
\end{equation}
where the $\beta>0$ is a hyperparameter used to balance the impact of the adversarial loss contribution. In Equation \ref{adv}, we use back propagation to optimize the prediction network $\Theta_F$ parameters with the total loss and then the adversarial network's $\Theta_A$ parameters using $H(S;\hat{S})$. Combining the impact of the two loss terms ensures that the prediction network will be penalized for producing an $R$ that can be used to reproduce the protected factor. The resulting prediction network should then be more resilient to bias with respect to the protected factor.

\begin{figure}[tb!]
    \centering
    %\vspace{-0.5cm}
    \includegraphics[width=\linewidth]{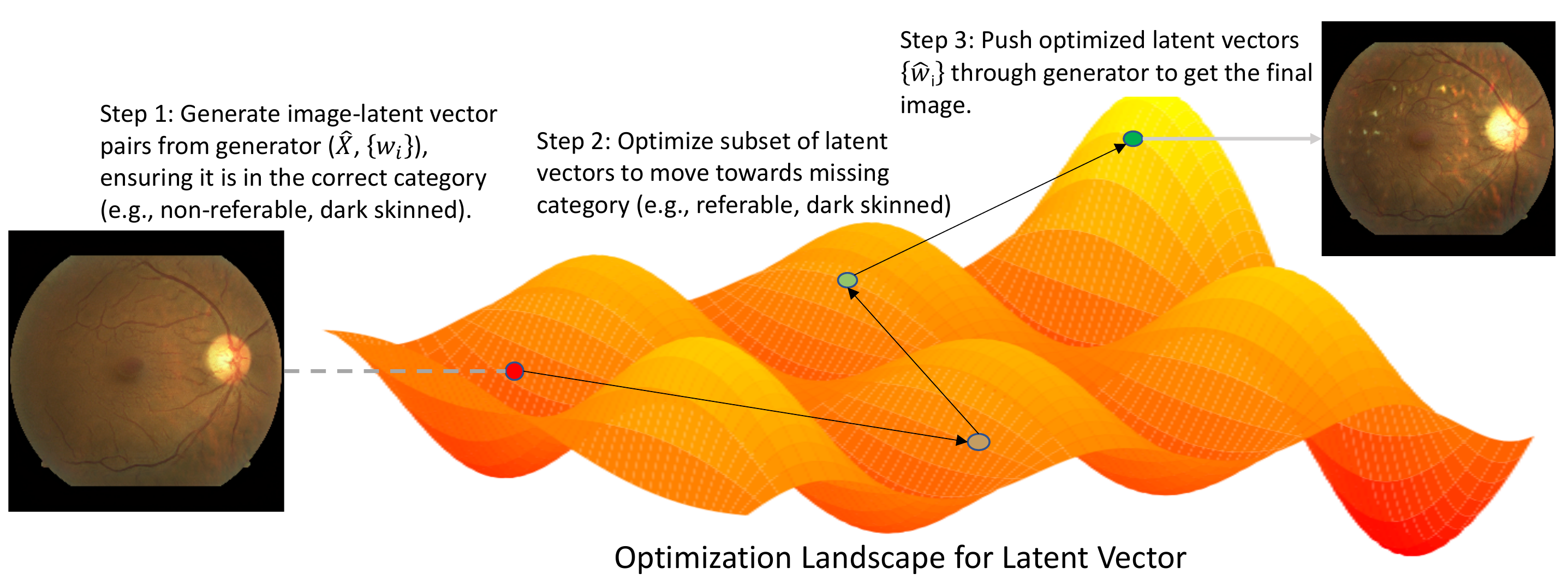}
    
    %\vspace{-3.5cm}
    \caption{Conceptual depiction of \IA{} which manipulates generated images to impart desired factors of underrepresented populations while keeping other factors invariant, shown on retinal data. Step 2 uses Equation \ref{gradupdate} to update the latent vectors.}
    \label{fig:latent_vector_steps}
    %\vspace{-0.5cm}
\end{figure}

\noindent\textbf{Intelligent Augmentation:} An alternative approach to improving model fairness is by generating more data for underrepresented populations to reduce dataset imbalance. For example, consider the retinal image analysis use case where the goal is to generate retinal images for underrepresented populations (e.g., dark skin individuals with referable diabetic retinopathy (DR)). These underrepresented retinal images could be generated from subpopulations whose data is more abundant (e.g., healthy dark skin individuals or DR-referable light skin individuals), while holding other image characteristics invariant (importantly, disease markers, but also vasculature and possibly other more subtle markers like gender markers~\cite{poplin2018prediction}). Our method does this through a combination of data generation using multiscale GANs (StyleGAN~\cite{karras2019style})  by controlling the direction of change in latent space via gradient descent that maximally alters a specific property of the image which can correspond to either a) the presence of the disease (i.e., $\hat{Y}=1$) or b) the presence of a specific attribute, for example ``dark skin individual'' (corresponding to $\hat{S}=1$). Once the GAN is trained on the available data, the desired transformation is obtained in three steps, as illustrated in Figure~\ref{fig:latent_vector_steps}: (1) sample from the generator to obtain image and style space vectors pairs ($\hat{X}$, $\{w_i\}$), where $i$ denotes the resolution scale; (2) map the vectors $\{w_i\}$ into $\{\hat{w}_{i}\}$, imparting the desired image property  change in latent space; (3) generate the corresponding image by mapping the latent space vector $\{\hat{w}_{i}\}$ from latent to image space (to get the final image with the factor changed). 
%StyleGAN, which includes a fully connected network that mapped a latent vector Z into a 512-length intermediate latent vector W that controls the generation and allows stylistic mixing of images (a so called 'style' vector). (This style vector W was used in StyleGAN to affect some factors of the image at the low scale coarse factors like skin tone and at the higher scale fine factors like hair).

Specifically, an image-space classifier $C_1$ is first trained on the property of interest on real images that are then used to label the vector-image pairs ($\hat{X}$, $\{w_i\}$), generated unconditionally from StyleGAN. To make the optimization process easier, we then train a second classifier $C_2$ that takes the $\hat{X}$'s corresponding style vectors $\{w_i\}$ as input and replicates the prediction of $C_1$ in the latent space. As a result, $C_2$'s gradient can then be applied to control $\{w_i\}$ directly, as gradient descent would yield a non-rectilinear trajectory in latent space that is maximally modifying with respect to (w.r.t.) the selected property. 

Consider, for example, a linear discriminator $C_2$ that is used to separate images in latent space $W$ w.r.t. the selected factor. This linear discriminator determines the hyperplane that separates training vectors into two classes: those that have vs. those that do not have the selected factor. Then the resulting direction normal to this hyperplane is maximally changing w.r.t. the selected factor. 

However, instead of a simple linear classifier we use a fully connected network for $C_2$. The loss function of this classifier is used to perform gradient descent in $W$ so as to arrive to a $\hat{W}$ which has the desired {\it softmax} value, thereby allowing fine control of the degree to which the factor is expressed in the image.

Mathematically, this is expressed as:
\begin{equation}
    w_{i, j+1} = w_{i, j} + \gamma \nabla_{w_{i,j}} \log (P(\hat{Y}=1|W_{i,j}=w_{i,j}), \hat{w}_{i} = w_{i, N}
    \label{gradupdate}
\end{equation} 

\noindent in case the property that is modified corresponds to a specific value of $Y$ (e.g., presence of disease) where $\gamma>0$ is a hyperparameter, $i$ is the resolution scale; $N$ is the total number of steps, $j$ is the current step to approach the desired factor,
and $P(\hat{Y}=1|W_{i,j}=w_{i,j})$
is the output of classifier $C_2$.
%and $C_2(\{w_{i,j}\})[1]$ is the probability value that $C_2$ outputs for $Y=1$. 
Alternatively, if we desire to impart a specific factor (e.g., accentuating dark skin or old age) to the image then we would use:
\begin{equation}
    w_{i, j+1} = w_{i, j} + \gamma \nabla_{w_{i,j}} \log (P(\hat{S}=1|W_{i,j}=w_{i,j}), \hat{w}_{i} = w_{i, N}
    \label{gradupdate2}
\end{equation}

\begin{figure}[tb!]
%\vspace{-0.3cm}
    \centering
    \includegraphics[width=\linewidth]{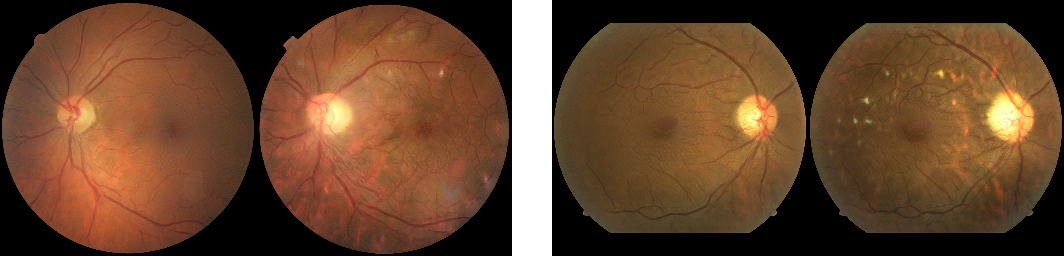}
    \caption{Examples of generating new data for underrepresented populations (here, dark skin individuals with referable diabetic retinopathy (DR)) via our \IA{}. Left are the original generated samples, and right are the transformed versions, characterized by DR lesions (bright spots) and variations in fundus background, see \cite{burlina2020addressing}.}
    \label{fig:eyepacs_trans}
    %\vspace{-0.5cm}
\end{figure}

\begin{figure}[tb!]
\centering
    
    \begin{subfigure}{0.48\textwidth}
    \centering
    
    \includegraphics[width=\linewidth]{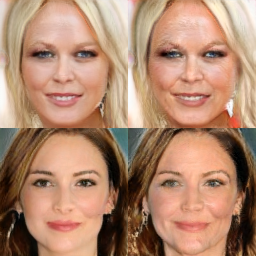}
    \caption{Young to Old}
    \label{Trans_Gender_CelebA}
    \end{subfigure}
    \begin{subfigure}{0.48\textwidth}
    \centering
    
    \includegraphics[width=\linewidth]{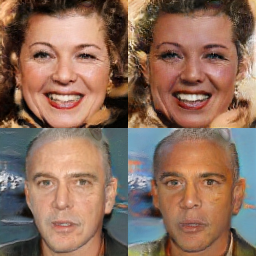}
    \caption{Light to Dark}
    \label{Trans_SkinColor_CelebA}
    \end{subfigure}
    \caption{Examples of new data generation for underrepresented populations (older females in (a), and older individuals with darker skin in (b)) via our \IA{} on faces. Left are the original generated samples, and right are the transformed versions.}
    \label{fig:celeba_trans}
    %\vspace{-0.5cm}
\end{figure}

Although we have a classifier that can tell the sensitive attribute from an image, this does not mean that other attributes are not affected. As StyleGAN encourages disentanglement between different resolution scales of the style vectors so that certain attributes are represented only in certain resolution scales, we only employ this update for specific scales to get $\{\hat{w}_{i}\}$, typically the finer styles as the coarse styles overly affect the image. We can then generate the final image by passing $\{\hat{w}_{i}\}$ through the generator. See Figures \ref{fig:eyepacs_trans} and \ref{fig:celeba_trans} for examples of this transformation.

\noindent\textbf{Training and Representation Alteration (\HI{}):}
The \IA{} method supplements underrepresented classes with additional samples, whereas the \AD{} method modifies the training procedures such that protected factors cannot be accurately predicted from learned representations produced by a classifier. Consequently, as these two methods affect complimentary domains, we combine the two methods into what we call the \HI{} method. 
Last, as the augmentation process approximates the true category that is missing in the dataset, any discrepancies between our synthetic category and images in the true category might be magnified by the application of \AD{}. Consequently, excluding synthetic images in the training of \AD{} can be beneficial. We denote this system by TARA+F and depict it in Figure~\ref{fig:TARA_network_diagram} via the deactivation of the teal dashed lines.

\section{Metrics}

In this section we recall existing metrics and propose novel metrics to characterize debiasing performance:

\noindent\textbf{Overall Accuracy and Accuracy Gap:}
The accuracy gap $acc_{gap}$ for a given model is measured as the difference in accuracy between the populations that have the maximum and the minimum accuracy. Reducing $acc_{gap}$ is a prime objective in conjunction with maintaining overall accuracy $acc$ for the debiased algorithm. The possible trade-off between both objectives of $acc$ and $acc_{gap}$ justifies the need for a single metric to assess a debiased models' performance.

\noindent\textbf{Minimum Accuracy ($acc_{min}$):}
A potential single performance metric $acc_{min}$ is based on the minimum accuracy across all protected subpopulations. Maximizing $acc_{min}$ follows the Rawlsian theory of distributive justice whereby the max-min fairness principle~\cite{rawls2001justice} suggests that maximizing the utility of the least advantaged subpopulation improves fairness in society~\cite{lahoti2020fairness}.

\noindent\textbf{Conjunctive Accuracy Improvement ($CAI_{\alpha}$):}
We propose two novel single performance measures as possible indicators of success. The first is a weighted linear combination of two differential terms including the (signed) accuracy gap decrease and the (signed) overall accuracy improvement,
where both terms are computed with respect to a baseline and candidate algorithm: 
\begin{equation}
CAI_{\alpha} = \alpha (acc_{gap}^b - acc_{gap}^d) + (1 - \alpha) (acc^d - acc^b) 
\end{equation}
where $\alpha$ is a weight coefficient and $acc^b$ and $acc^d$ denote the accuracy of the baseline and debiased models, respectively. Similarly, $gap^b$ and $gap^d$ represent the accuracy gap of the baseline and debiased models, respectively. We call this metric the Conjunctive Accuracy Improvement ($CAI_{\alpha}$). Deciding how to weigh the respective importance of the two metrics is a matter beyond engineering which also should involve ethicists and policy makers (see Section~\ref{sec:Discussion} for details). The cases of $\alpha=0.5$ and $0.75$ are reported here for illustrative purposes to motivate future discussions.

\noindent\textbf{Generalizing to other metrics ($CAUCI_{\alpha}$):}
The second proposed metric extends $CAI_{\alpha}$ to AUC, which we call the Compound AUC Improvement (abbreviated henceforth as $CAUCI_{\alpha}$). We also use AUC gap and minimum AUC for consistency. These ideas can similarly be extended to other metrics such as F1-score (not pursued here).

\section{Datasets Used and Individual Typology Angle}
\subsection{OSMI Mental Health}

First, to understand the effect of the proposed novel metrics, we used~\AD{} (without~\IA{}) on OSMI (OSMI Mental Health in Tech Survey 2016~\cite{KaggleOSMIdataset}). This tabular records dataset released on Kaggle to encourage evaluation of the state of mental health across the technology industry. We investigated gender and age bias when predicting whether a person sought treatment for mental illness. We omitted the {\it other} class from the three possible gender classes ({\it male, female, other}) due to ambiguous sample quantity and quality. For age debiasing we simplified the problem to a binary class: younger ($\le 40$ years old) and older ($> 40$ years old). Eight tabular features were used to train the binary prediction network. 

\subsection{Individual Typology Angle} 

Next we describe image datasets used with all (\AD{},~\IA{}, and~\HI{}) debiasing methods. Before doing so, we introduce a method used as a proxy for race and skin tone\footnote{Our usage of the terms {\em race}, {\em ethnicity} and {\em skin tone} are consistent with~\cite{burlina2020addressing} and \cite{christiansen2020ama}.} protected factor, the Individual Topology Angle (ITA)~\cite{wilkes2015fitzpatrick}. The ITA was found to correlate with the Fitzpatrick Skin Type typically used in dermatology for characterizing the skin color of an individual. To compute ITA, an image is first converted to the CIELab color space, which was designed to match perceptual differences with differences in numerical values $L$ of lightness, scale $a$ between red and green, and scale $b$ between blue and yellow. The computed per pixel ITA is:
\begin{equation}
    ITA = \frac{180}{\pi} \arctan \left(\frac{L - 50}{b}\right).
    \label{eq:ITA}
\end{equation}
The ITA for a given image was then computed by averaging pixel values over some masked area, which is determined for each dataset (see Appendix). 

\begin{figure}
    \centering
    \begin{subfigure}{0.49\textwidth}
    \includegraphics[width=\linewidth]{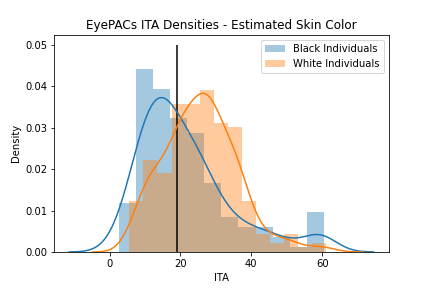}
    \caption{EyePACs ITA Density}
    \label{ITA_race_EyePacs}
    \end{subfigure}
    \begin{subfigure}{0.45\textwidth}
    \includegraphics[width=\linewidth]{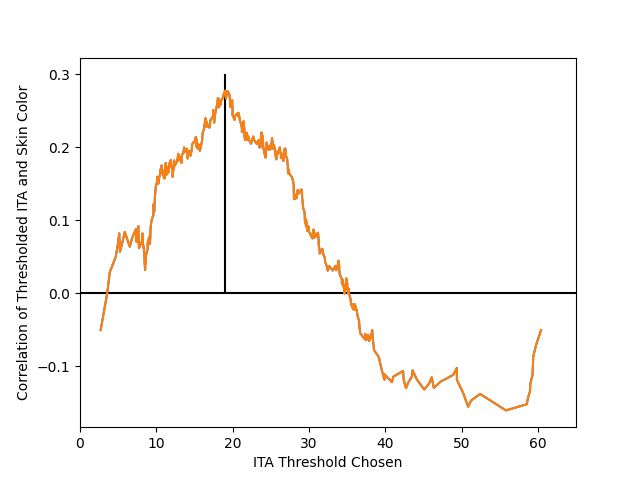}
    \caption{Correlations between ITA and S.}
    \label{ITA_corr_Eyepacs}
    \end{subfigure}
    \caption{We use the ITA as a proxy for skin color. The ITA distribution is shown for retinal (EyePACS). Black line denotes the cutoff used to create a binary variable as a sensitive factor. Both figures show how well the cutoff aligns with the true sensitive attribute, with the figure on the left showing that the correlation remains strong even in a local neighborhood of the value chosen.}
    %\vspace{-0.5cm}
\end{figure}

\subsection{EyePACs} 

Sourced from the Kaggle Diabetic Retinopathy challenge, EyePACS~(\cite{KaggleEyePACSdataset}) includes retinal fundus images of individuals potentially affected by diabetic retinopathy (DR). The original labels, ranging from 0 (not afflicted) to 4 (most severe), were binarized to denote a status greater than (mild DR). In \cite{burlina2020addressing}, the dataset was also annotated by a clinician with an additional label that reflects an estimation of the individual's skin color in relation to their race: a binary factor reflecting image markers related to race such as darker pigmentation in the fundus, thicker blood vessels, larger cup to optical disk sizes usually associated with Black individuals rather than White individuals.

Although we have annotations for this sensitive attribute, the data collection was expensive, time consuming, and incomplete, covering only a small portion of the dataset that was then extrapolated to the rest of the dataset. There are issues with uncertainty vis-a-vis self-reporting of race.
Consequently, rather than training using this attribute as $S$, we instead investigate using another attribute as $S$ that requires no manual annotation, namely the ITA, and evaluate the resulting systems on both types of sensitive attributes, using ITA as a proxy for the estimated skin color.

To calculate the ITA, we added another binary variable denoting  fundi with ITA $\leq 19$, which is taken to mean dark skinned. This cutoff is chosen to match established categories in~\cite{kinyanjui2020fairness} and to mimic the previous label for the annotated race, as the fundus color should be a major component of that label. From Figure~\ref{ITA_race_EyePacs}, we see that the cutoff closely separates the two distributions induced by the race label, but not exactly due to potential existence of other criteria in the labeled race (optical disk size) as well as retinal artifacts affecting computing the ITA. We also see from Figure~\ref{ITA_corr_Eyepacs} that this cutoff maximizes the correlation between the thresholded ITAs and the true sensitive factor, though the correlation remains near the same levels or decreases gracefully in a local neighborhood from the ITA equaling 15 to 24 which suggests some insensitivity property.

\subsection{CelebA}
\begin{figure*}
    \centering
    \includegraphics[width=0.5\linewidth]{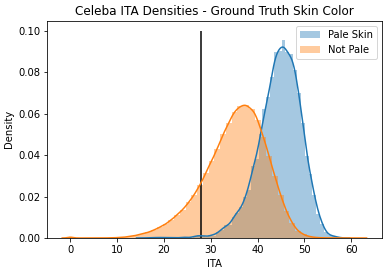}
    \caption{Use of ITA for skin color in facial imagery. Black line denotes cutoff for sensitive factor, which is consistent with the label for pale skin and established skin categories.}
    \label{ITA_SkinColor_CelebA}
\end{figure*}
CelebA consists over 200,000 celebrity faces with various descriptive factors, including gender, considered as a protected factor; and age, as a prediction target. As with EyePACs, ITA is used as a protected factor (as proxy for skin color), and annotated for each image in a manner similar to \cite{merler2019diversity}. Figure~\ref{ITA_SkinColor_CelebA} plots the distribution of ITA computed for CelebA vs.\ ground truth labels for the existence of pale skin. This figure demonstrates separation between the two modes for those with pale skin vs. those without pale skin, with some overlap between ITA of 30 and 50. One potential factor that may hurt separation is a significant number of celebrities having bright lights shining directly on them, overly lightening their skin relative to their actual skin color. To focus on extreme cases of data bias, we use a threshold of 28, which matches category thresholds from \cite{kinyanjui2020fairness} and also excludes almost all individuals denoted as Pale Skin.

\section{Experiments}\label{sec:Experiments}

\noindent\textbf{Domain Generalization:} We address the use case of pronounced data imbalance, where we constructed training partitions in a manner where a certain category of targeted and protected factors, namely the ones that contain the fewest members, were excluded entirely. This tested how well the proposed method generalized to unseen categories, a bias problem akin to {\it domain generalization}. Alternative and more benign cases were reported in the Appendix, in addition to extensive datasets details.

\noindent\textbf{ Post-processing via Best Thresholding:} Alongside the other methods we discussed previously, we also evaluate against a baseline algorithm that consisted of changing the threshold used for prediction via the method in \cite{hardt2016}. This post-processing method (henceforth referred to as \BT{} or ``Best Thre.'' for short) takes in a baseline classifier and validation dataset. It tunes the decision rule thresholds for each sensitive population such that the true positive rate and false positive rate match for all subpopulations. 

As data in the missing category now becomes necessary to have non-trivial solutions to implement this \BT{} algorithm, we use a separate validation set balanced across both the sensitive attribute and the target attribute for this method which is not used by any other method; so this method denotes the best possible choice for thresholding for the baseline classifier. Since this baseline uses significantly more information, the comparison with our proposed methods is not entirely equitable.
\begin{figure}[thb!]
    \centering
    \includegraphics[width=0.7\textwidth]{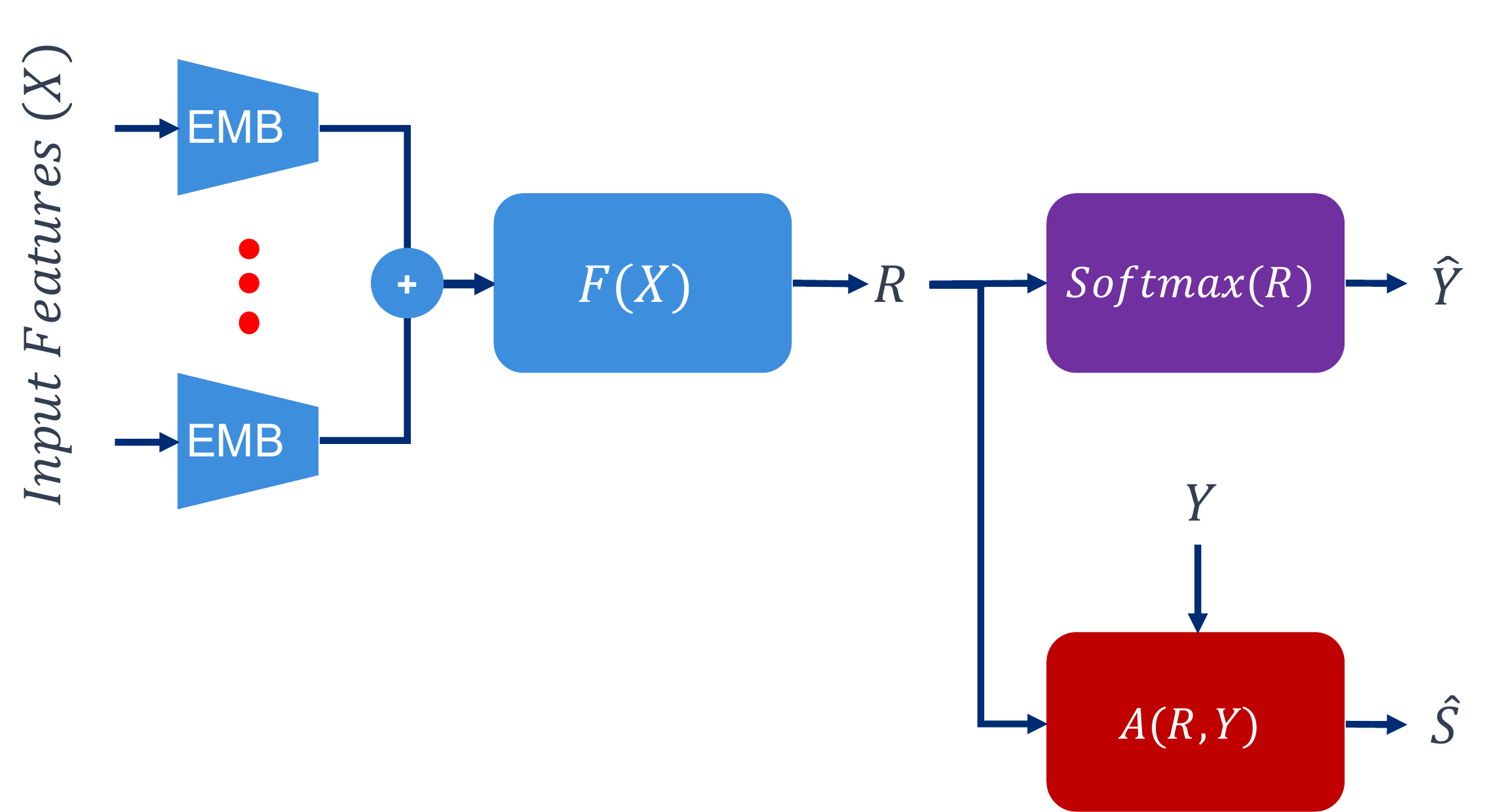}
    \vspace{-0.25cm}
    \caption{Adversarial Independence Neural Network ($EMB+AD$) data using an embedding-based prediction module ($F$), for a given tabular input feature ($X$), to produce an internal representation ($R$) and corresponding prediction ($\hat{Y}$).}
    \label{fig:adv_debias_network_diagram}
\end{figure}
%\vspace{-0.25cm}

\begin{table}[thb!]
\scriptsize
    \begin{center}
    
\caption{Performance metrics for debiasing methods on OSMI predicting $Y$= sought mental health treatment, trained on partitioning with respect to $S$ = Gender, and evaluated on a test set balanced across treatment status and gender (M/F). Methods include: embeddings prediction network (EMB), noise debias (Noise), adversarial debias (AD). 95\% confidence intervals are shown in parentheses.} \label{tbl:2016_balanced_results_gender}
        \begin{tabular}{l||c|c|c}
        \toprule

            Metrics & Baseline (EMB) & EMB+Noise & EMB+AD \\

        \hline
        \midrule
            $acc (\%)$ & 63.2 (5.1) & 67.8 (4.9) & \textbf{82.2 (4.0)} \\
            \hline
            
            $acc_{gap} (\%)$ & 20.7 (0.9) & 24.14 (1.4) & \textbf{4.59 (0.6)} \\
            \hline
            
            $acc_{min} (\%)$ (subpop.) & 52.9 (F) & 55.8 (F) & \textbf{79.9 (M)} \\
            \hline
            
            $CAI_{0.5} (\%)$ & - & 0.6 & \textbf{17.5} \\
            \hline
            
            $CAI_{0.75} (\%)$ & - & -1.4 & \textbf{16.8} \\
            \hline
            \midrule
            
            $AUC$ & 0.721 (0.047) & 0.803 (0.042) & \textbf{0.859 (0.037)} \\
            \hline
            
            $AUC_{gap}$ & \textbf{0.061 (0.009)} & 0.061 (0.010) & 0.081 (0.013) \\
            \hline
            
            $AUC_{min}$ (subpop.) & 0.817 (M) & \textbf{0.843 (M)} & 0.824 (M) \\
            \hline
           
            $CAUCI_{0.5}$ & - & 0.041 & \textbf{0.059} \\
            \hline
            
            $CAUCI_{0.75}$ & - & \textbf{0.020} & 0.020 \\
        \bottomrule
        \end{tabular}
    \end{center}
    %\vspace{-0.6cm}
\end{table}

\begin{table}[tbh!]
\scriptsize
    \begin{center}
\caption{Performance metrics for debiasing methods on OSMI predicting $Y$= sought mental health treatment, trained on partitioning with respect to $S=$ Age, and evaluated on a test set balanced across treatment status and age (Older/Younger).} \label{tbl:2016_balanced_results_age}
        \begin{tabular}{l||c|c|c}
        \toprule

            Metrics & Baseline (EMB) & EMB+Noise & EMB+AD \\

        \hline
        \midrule
            
            $acc$ & 68.4 (4.3) & 75.0 (4.0) & \textbf{80.8 (3.6)} \\
            \hline
            
            $acc_{gap}$ & 21.7 (1.2) & 18.1 (1.4) & \textbf{1.3 (0.1)} \\
            \hline
            
            $acc_{min}$ & 57.5 (Older) & 65.9 (Older) & \textbf{80.1 (Older)} \\
            \hline
            
            $CAI_{0.5}$ & - & 5.09 & \textbf{16.37} \\
            \hline
            
            $CAI_{0.75}$ & - & 4.32 & \textbf{18.36} \\
            \hline
            \midrule
            
            $AUC$ & 0.766 (0.044) & 0.821 (0.035) & \textbf{0.879 (0.030)} \\
            \hline

            $AUC_{gap}$ & 0.001 (0.0002) & \textbf{0.0003 (0.0001)} & 0.014 (0.002) \\
            \hline

            $AUC_{min}$ & 0.865 (Older) & \textbf{0.878 (Older)} & 0.876 (Younger) \\
            \hline

            $CAUCI_{0.5}$ & - & 0.028 & \textbf{0.050} \\
            \hline
            
            $CAUCI_{0.75}$ & - & 0.015 & \textbf{0.019} \\
        \bottomrule
        \end{tabular}
    \end{center}
    %\vspace{-0.6cm}
\end{table}

\noindent\textbf{OSMI Experiments:}
We developed a series of experiments in order to evaluate the impact of \AD{} on improving fairness, with respect to gender and age, in an extreme case of domain generalization when predicting whether a given person sought mental health treatment using the OSMI dataset, and assessing the novel performance metrics before moving to image data. Due to the tabular nature of the dataset we did not perform any \IA{} and instead explored \AD{} more extensively. In implementing \AD{} on OSMI, we provided the logits to the adversary due to simplicity. Our proposed \AD{} debiasing approach ({\it EMB+AD}, depicted in Figure~\ref{fig:adv_debias_network_diagram}) included a prediction network with an embedding for each of input features concatenated and forwarded to a fully connected layer ({\it EMB}), while the adversarial debiasing network was a single fully connected layer ({\it AD}). The gender debiasing results shown in Table \ref{tbl:2016_balanced_results_gender} reflect a 19\% increase in overall accuracy and 16.1\% reduction in accuracy gap and between male and female when using {\it EMB+AD} compared with the baseline method {\it EMB}. The $CAI_{\alpha}$ mirrored both of these accuracy improvements with $CAI_{0.5}$ and $CAI_{0.75}$ being the highest for {\it EMB+AD}. Similar to gender debiasing, the age debiasing results in Table~\ref{tbl:2016_balanced_results_age} suggest {\it EMB+AD} had superior performance over the other two methods. Moreover, applying Gaussian noise ({\it Noise}) to the prediction module logits as regularization was not found to be as beneficial as using {\it AD} to explicitly reduce bias towards the protected factor. Results also illustrated the effectiveness of the conjunctive metrics in reflecting the best overall fairness performance in a compact manner.

\noindent\textbf{Implementation Details for All Image Experiments:}  We use ResNet50 as a baseline classifier and also as base architecture in our \AD{} image experiments.  For image data, since \AD{} using the logits performed close to the baseline in most respects and did not affect the model, we used the activations prior to the final linear layer as the input to the adversary. We also used two settings for $\beta=0.5,1.0$ and reported the best overall accuracy. 

\noindent\textbf{EyePACS Experiments:}
We tested bias induced by domain generalization, by excluding DR-referable fundi from dark skin individuals from training, and kept the training dataset otherwise balanced across the DR label. Testing used two cases. The first uses a test set equally balanced between the disease status (DR) and the skin color (ITA), with 600 examples per category. The second case tests a type of generalization of the proxy protected factor used in training and testing/inference; i.e. while we use ITA to debias the model against at training time, we now do testing by using actual race labels as protected factor instead of ITA. This was implemented done by matching the test set conditions used in \cite{burlina2020addressing}, that were equally balanced across disease status and estimated skin color by the clinician, with 100 examples per category used. The two test sets were disjoint from each other.
Table~\ref{ITA_DR} shows that \HI{}+F performed best in terms of $CAI_{0.5}$, with  accuracy gap reduced to almost zero while overall accuracy increased by 7.5\%. For the second test case in Table~\ref{AA_DR}, \HI{}+F performed best, and reduced the accuracy gap to near zero. In both cases, \HI{} outperformed \AD{} which in turn outperformed~\IA{}, with all methods beating the baseline. 
\begin{table}[tbh!]
\scriptsize
    \begin{center}
 
    \caption{Performance metrics for debiasing methods on EyePACs predicting $Y$= DR Status, trained on partitioning with respect to $S=$ ITA, and evaluated on a test set balanced across DR status and ITA. Methods include baseline, IA: \IA{}, AD: \AD{}, TARA: \HI{} with or without filtering (F). * denotes the method which requires additional information outside of what other methods are allowed, and numbers in bold are the best for the metric between methods using equal amounts of data. The value in parenthesises in the first row is the $\beta$ used for AD, other numerical parenthesises denote error margins, and the parenthesises with text denote the subpopulation that corresponds to the minimum accuracy/AUC.}
    \begin{tabular}{l||c|c|c|c|c|c}
    \toprule
        Metrics & Baseline (0.0) & Best Thre. (0.0)* & AD (0.5) & IA (0.0) & TARA (0.5)  & TARA+F (1.0) \\
        \hline
        
        \midrule
        $acc$ & 70.0 (1.8) & 71.5 (1.8) & 76.1 (1.7) & 71.5 (1.8) & \textbf{78.0 (1.7)} & 77.5 (1.7) \\
        \hline
        
        $acc_{gap}$ & 3.5 (3.7) & 7.2 (3.8) & 2.41 (3.4) & 1.5 (3.6) & 2.34 (3.3) & \textbf{0.16 (3.3)} \\
        \hline
        
        $acc_{min}$ & 68.3 (Dark) & 67.8 (Light) & 74.9 (Light) & 70.8 (Dark) & 76.8 (Light) & \textbf{77.4 (Dark)} \\
        \hline
        
        $CAI_{0.5}$ & - & -1.1 & 3.6 & 1.8 & 4.6 & \textbf{5.4} \\
        \hline
        
        $CAI_{0.75}$ & - & -2.4 & 2.4 & 1.9 & 2.9 & \textbf{4.4} \\
        \hline
        \midrule
        
        $AUC$ & 0.786 (0.016) & 0.786 (0.016) & 0.835 (0.015) & 0.773 (0.017) & 0.851 (0.014)  & \textbf{0.855 (0.014)} \\
        \hline
        
        $AUC_{gap}$ & 0.032 (0.031) & 0.032 (0.031) & 0.030 (0.029) & \textbf{0.005 (0.032)} & 0.019 (0.028) & 0.031 (0.028) \\
        \hline
        
        $AUC_{min}$  & 0.794 (Light) & 0.794 (Light) & 0.832 (Light) & 0.797 (Light) & 0.848 (Light) &  \textbf{0.847 (Light)} \\
        \hline
        
        $CAUCI_{0.5}$ & - & 0.0 & 0.025 & 0.007 & \textbf{0.039} & 0.035 \\
        \hline
        
        $CAUCI_{0.75}$ & - & 0.0 & 0.014 & 0.017 & \textbf{0.026} & 0.018 \\
\bottomrule

    \end{tabular}
           
    \label{ITA_DR}
    \end{center}
\end{table}
\begin{table}[]
\tiny
    \centering
    \caption{Performance metrics for debiasing methods on EyePACs predicting $Y$= DR Status, trained on partitioning with respect to $S=$ ITA, and evaluated on a test set balanced across DR status and estimated skin color/race. The last two columns compare with methods in~\cite{burlina2020addressing} which were trained with the partitioning with $s=$ race. * denotes the method which requires additional information outside of what other methods are allowed.}
    \begin{tabular}{l||c|c|c|c|c|c || c | c}
    \toprule
        Metrics & Baseline (0.0) & Best Thre. (0.0)* & AD (0.5) & IA (0.0) & TARA (0.5)  & TARA+F (1.0) & (Pr.) Baseline & (Pr.) IA\\
        \hline
        
        \midrule
        $acc$ & 71.8 (4.4) & 69.8 (4.4) & 76.0 (4.2) & 73.3 (4.3) & 76.5 (4.2)  & \textbf{78.8 (4.0)} & 66.8 (4.6) & 74.8 (4.3) \\
        \hline
        
        $acc_{gap}$ & 10.5 (8.8) & 3.5 (8.3) & 5.0 (8.5) & 7.5 (8.6) & 4.0 (8.3) & \textbf{0.5 (8.0)} & 12.5 (9.2) & 7.5 (8.5) \\
        \hline
        
        $acc_{min}$ & 66.5 (Black) & 68.0 (Black) & 73.5 (Black) & 69.5 (Black) & 74.5 (Black) & \textbf{78.5 (Black)} &  60.5 (Black) & 71.0 (Black) \\
        \hline
        
        $CAI_{0.5}$ & - & 2.5 & 4.9 & 2.3 & 5.6 & \textbf{8.5} & - & - \\
        \hline
        
        $CAI_{0.75}$ & - & 4.8 & 5.2 & 2.6 & 6.1 & \textbf{9.3} & - & - \\
        \hline
        \midrule
        
        $AUC$ & 0.771 (0.041)& 0.771 (0.041) & 0.837 (0.036) & 0.794 (0.040) & 0.861 (0.034) & \textbf{0.870 (0.033)} & - & - \\
        \hline
        
        $AUC_{gap}$ & 0.123 (0.081)& 0.123 (0.081) & 0.045 (0.073) & 0.055 (0.078) & 0.030 (0.068) & \textbf{0.011 (0.066)} & - & - \\
        \hline
        
        $AUC_{min}$ & 0.711 (Black) & 0.711 (Black) & 0.813 (Black) & 0.770 (Black) & 0.846 (Black) & \textbf{0.866 (Black)} & - & - \\
        \hline
        
        $CAUCI_{0.5}$ & - & 0 & 0.072 & 0.046 & 0.092 & \textbf{0.106} & - & - \\
        \hline
        $CAUCI_{0.75}$ & - & 0 & 0.075 & 0.057 & 0.092 & \textbf{0.109} & - & - \\
\bottomrule

    \end{tabular}
    \label{AA_DR}
\end{table}

\begin{table}[!tbh]
\scriptsize
    \centering
    \caption{Performance metrics for debiasing methods on CelebA predicting $Y$= Age, trained on partitioning with respect to $S=$ Gender, and evaluated on a test set balanced across age and gender. * denotes the method which requires additional information outside of what other methods are allowed.}
    
    \begin{tabular}{l||c|c|c|c|c|c}
    \toprule
        Metrics & Baseline (0.0) & Best Thre. (0.0)* & AD (1.0) & IA (0.0) & TARA (0.5) &  TARA+F (1.0)\\
        \hline
        
        \midrule
        $acc$ & 69.1 (1.0) & 70.0 (1.0) & 73.3 (1.0)  & 69.9 (1.0) & 72.09 (1.0) &  \textbf{73.7 (1.0)} \\
        \hline
        
        $acc_{gap}$ & 21.7 (2.0) & 0.7 (1.4) & 14.0 (1.9) & 13.1 (2.0) & 18.4 (1.9) & \textbf{11.8 (1.9)} \\
        \hline
        
        $acc_{min}$ & 58.3 (F) & 69.3 (F) & 66.3 (F) & 63.4 (F) & 62.9 (F) & \textbf{67.8 (F)} \\
        \hline
        
        $CAI_{0.5}$ & - &10.0 & 5.9 & 4.7  & 3.2 & \textbf{7.3} \\
        \hline
        
        $CAI_{0.75}$ & - &15.5 & 6.8 & 6.6 & 3.2 & \textbf{8.5} \\
        \hline
        \midrule
        
        $AUC$ & 0.751 (0.010) & 0.751 (0.010) & \textbf{0.821 (0.008)} & 0.766 (0.009) & 0.804 (0.009) & 0.816 (0.009) \\
        \hline
        
        $AUC_{gap}$ & 0.124 (0.016)  & 0.124 (0.016) & 0.104 (0.015) & 0.123  (0.016) & 0.128 (0.016) & \textbf{0.101 (0.016)}\\
        \hline
        
        $AUC_{min}$ & 0.772 (F) & 0.772 (F) & \textbf{0.799 (F)} & 0.763 (F) &  0.773 (F) & 0.775 (F)\\
        \hline
        
        $CAUCI_{0.5}$ & - & 0.0 &  \textbf{0.045} & 0.008 &  0.025  & \textbf{0.045}\\
        \hline
        
        $CAUCI_{0.75}$ & - & 0.0 & 0.033 & 0.005 &   0.010  & \textbf{0.034}\\
\bottomrule

    \end{tabular}
    \label{table:CelebA_Gender_Age}
\end{table}

\begin{table}[!tbh]
\scriptsize
    \centering
    \caption{Performance metrics for debiasing methods on CelebA predicting $Y$= Age, trained on partitioning with respect to $S=$ Skin Color, and evaluated on a test set balanced across age and skin color. * denotes the method which requires additional information outside of what other methods are allowed.}
    \begin{tabular}{l||c|c|c|c|c|c}
    \toprule
        Metrics & Baseline (0.0) & Best Thre. (0.0)* & AD (0.5) & IA (0.0) & TARA (0.5) & TARA+F (1.0) \\
        \hline
        
        \midrule
        $acc$ & 74.4 (1.0) & 73.2 &  \textbf{76.5 (0.9)} & 75.29 (1.0) &  69.58 (1.0) & 75.05 (1.0)\\
        \hline
        
        $acc_{gap}$ & 13.9 (1.9) & 1.0 (1.5) & 9.6 (1.9) & 9.18 (1.9) & 12.15 (2.0) & \textbf{7.25 (1.9)}  \\
        \hline
        
        $acc_{min}$ & 67.5 (Dark) & 72.7 (White) & \textbf{71.7 (Dark)} & 70.7 (Dark) & 63.5 (Dark) & 71.4 (Dark) \\
        \hline
        
        $CAI_{0.5}$ & - & 5.9 & 3.2 & 2.8  &  -1.5 & \textbf{3.6}\\
        \hline
        
        $CAI_{0.75}$ & - & 10.1 & 3.8 & 3.8 & 0.1 & \textbf{5.2} \\
        \hline
        \midrule
        
        $AUC$ & 0.818 (0.008)  & 0.818 (0.008) & \textbf{0.861  (0.008)} & 0.828 (0.008) & 0.806 (0.009)   & 0.845 (0.008) \\
        \hline
        
        $AUC_{gap}$ & 0.100 (0.015)  & 0.100 (0.015) & \textbf{0.056 (0.015)} & 0.071 (0.015) & 0.068 (0.017)  &  0.066 (0.016)\\
        \hline
        
        $AUC_{min}$ & 0.804 (Dark)  & 0.804 (Dark) & \textbf{0.846 (Dark)} & 0.820 (Dark) & 0.788 (Dark) & 0.813 (Dark) \\
        \hline
        
        $CAUCI_{0.5}$ & - & 0.0 & \textbf{0.044} & 0.019 & 0.010 & 0.030 \\
        \hline
        
        $CAUCI_{0.75}$ & - & 0.0 & \textbf{0.044} & 0.024 & 0.021  &  0.032 \\
\bottomrule

    \end{tabular}
    \label{table:CelebA_Skincolor_Age}
\end{table}

\noindent\textbf{CelebA Experiments:} For CelebA, we conducted two experiments of partitioning to predict age for each protected factor (gender and skin color). For each protected factor we proceeded similarly as with EyePACS, in that we excluded the smallest subpopulation from our training dataset, older females for gender partitioning and older dark skinned individuals for skin color partitioning. The test set used for both experiments was balanced across age, gender, and skin color with 1000 examples per category.
From Table~\ref{table:CelebA_Gender_Age}, we observe the following behavior when trying to reduce CelebA gender bias: setting aside the \BT{} method, ---which outperforms other methods in terms of accuracy-based metrics but uses extra data--- \HI{}+F outperforms all comparable (same amount of data) methods. In particular, the \AD{} and \HI{}+F approaches have the best performance (even overtaking the \BT{}) when considering the AUC-based metrics.

With regards to minimizing CelebA skin color bias, shown in Table~\ref{table:CelebA_Skincolor_Age}, we note that \HI{} performed worse than the baseline in terms of overall accuracy, with only a slight decrease in the accuracy gap (as reflected by the $CAI$ scores). However, \HI{}+F performed best in terms of accuracy gap and $CAI$. Again, the above comparisons do not include the \BT{} baseline in order to ensure that the same quantity of data was used by all compared methods.
Overall, the CelebA results echo the OSMI and EyePACS experiments in demonstrating the benefit of the conjunctive metrics in identifying the debiasing methods that exhibit the best overall fairness performance (in this case, achieved by~\HI{}+F).
% Finally, in Table~\ref{table:CelebA_Combo_Age}, we see that IA performs slightly worse at the accuracy gap and minimum accuracy. However, the total accuracy is up due to every other subpopulation (Light skinned males, light skinned females, dark skinned males) increasing in accuracy.
\section{Discussion}
\label{sec:Discussion}

\noindent\textbf{Experimental Results:} 
Leaving aside Best Thresholding for fair comparison, all methods tested, except \HI{} with no filtering in Table~\ref{table:CelebA_Skincolor_Age}, successfully show improvements in both overall accuracy and accuracy gap compared to the baseline as summarized by $CAI_{0.5}$. \HI{} with filtering in Table~\ref{AA_DR} was the best over  all metrics for EyePACs, with \HI{} with no filtering coming in second and \AD{} in third.  For both CelebA experiments, \AD{} and \HI{} with filtering were competitive with each other as the best of all methods, with the former method having better overall accuracy for skin color partitioning and consistently better AUC metrics for both partitions. The latter had a better accuracy gap for both partitions and a better accuracy overall for gender partitioning. 

When using the Best Thresholding  method with extra data not allowed otherwise in any other methods in our setting, we see improvements compared to the baseline for both EyePACs and CelebA regarding fairness. This is likely due to the post-processing strictly enforcing equalized odds on the validation dataset. Although this best threshold method primarily showcases the capacity of the baseline to elicit fair decisions rather than act as a direct comparison, this post-processing could be applied to our methods as well (and will be done as future work). Moreover, as this tuning utilizes the ROC curve, improvements in the various AUC metrics for our methods imply tuning may have as beneficial of an effect for accuracy metrics as on the baseline classifier.

%Moreover, while it is possible to maximize fairness and accuracy simultaneously it is also possible to have a model that accounts for social selection bias but leads to poor performance for all subpopulations.

For all image experiments, \IA{} was outperformed by \AD{} and at least one variant of \HI{}, though it still had improvements in the accuracy gap even when compared to \AD{}. However, \AD{} was sensitive to $\beta$, as using $\beta = 1.0$ for both \AD{} and the non-filtering \HI{} resulted in overall accuracies of approximately $60\%$ with similar accuracy gaps to the current implementation. 
Note that our \AD{} method is similar to~\cite{zhang2018mitigating} with the exception that we draw the representation for the second to last layer of the network, using the output of the flatten layer in ResNet50. The use of softmax as in~\cite{zhang2018mitigating} did not confer any benefits over the baseline. These results were not reported in our experiments. 

In comparison to \AD{}, \IA{} only affects the dataset, training was less complicated, and could extend to other domains, such as segmentation, an avenue we leave for future work. Now while \IA{} can be regarded as a pre-processing method that augments the training dataset and \AD{} modifies the training process, including the best threshold technique fills out the taxonomy of possible modifications to the overall process, though it has similar issues to oversampling the minority subpopulation as it requires data from the missing category to successfully carry out the method.

When we debias with respect to ITA as a sensitive factor, we note from Table~\ref{ITA_DR} that the resulting system succeeds in achieving parity for ITA.
Furthermore, we note that this system succeeds at achieving parity with regard to another sensitive factor (i.e., race) for which ITA was a proxy; see Table~\ref{AA_DR}. 

This suggests that the fundus pigmentation, out of all the image markers that comprise the presumed race, is the principal marker causing bias, and that ITA, which did not require a specialist to acquire, can be used as a surrogate factor for the difficult to acquire demographic information. 
In sum, we posit that ITA is a robust proxy measure for the debiasing method with race as a sensitive factor for EyePACS. Future work could include a closer examination of robustness of debiasing methods to mismatch in sensitive factors.

Note, our construction for domain generalization does not hold in Table~\ref{AA_DR} as some Black referable individuals were present in the training set. This may explain why the baseline accuracy was higher than the baseline from~\cite{burlina2020addressing}, though it was a different training dataset. On the other hand, the factor omitted between each CelebA experiment had varying effect on the baseline's overall accuracy on the shared test set, with the omission of older females having a larger impact on the baseline for the accuracy gap and overall accuracy compared to excluding older darker individuals. 

For the tabular data, we performed a set of experiments on the OSMI dataset to understand the network configuration that maximized fairness under our generalization experiment. The \AD{} method ({\it EMB+AD}) was found to maximize overall accuracy, $CAI_{0.5}$, $CAUCI_{0.5}$, and minimum accuracy of the subpopulations that were under represented for both protected factors (female and old). {\it EMB+AD} was also shown to target debiasing of the specific factors instead of acting as general regularization like the {\it Noise} debiasing approach. The overall results for gender and age debiasing suggest that using adversarial learning to reduce the prediction network's bias towards a protected class can reduce bias and encourage it to identify more optimal network weights. A more extensive ablation study can be found in the Appendix.

The test sets for each of the datasets were carefully balanced with respect to the given labels $Y$ and protected factors $S$, while also including a completely excluded subpopulation. The training sets, on the other hand, had natural selection bias inherent in the dataset and contained no samples from the excluded subpopulation. As a result, the distributions of the train and test sets differed with respect to $Y$ and $S$. 
%However, for each of these datasets $Y$ should have been independent of $S$, which suggested that enforcing fairness across $S$ should have {\it increased} classification accuracy because independence was being enforced~\cite{Wick2019NEURIPS}. This was further reinforced by the empirical results we presented on each of the datasets, which showed accuracy improvements for methods that enforced independence between $Y$ and $S$, such as \AD{} and TARA.

Generally a tradeoff may emerge between utility (overall accuracy) and fairness. Interestingly, in some situations it may be possible to increase accuracy and maximize fairness simultaneously by accounting for dataset label $Y$ and/or training set selection bias, as argued in~\cite{Wick2019NEURIPS}. Training data bias can exist that is not directly attributable to the label or selection bias of the data construction or acquisition process, but rather the biased data can be a by-product of societal selection bias. For example, celebrity-based image datasets, such as CelebA, contain more images of light skin than dark skin celebrities because it is a reflection of the inherent demographics of Hollywood celebrities, which can be attributed to many factors (e.g., opportunity, interest, etc.) which we call societal selection bias. These demographics result in age distributions that do not match across skin tone, despite logic dictating that skin tone (protected factor) should be independent of the age of a celebrity (task label) in a picture. As a result, a model would need to learn to account for the societal selection bias in order to fully account for bias in datasets such as CelebA. In cases where accounting for societal selection bias of a protected factor is particularly challenging, a trade-off will inevitably appear between fairness and accuracy, which results in an overall performance that can be quantified by our proposed $CAI_{\alpha}$ and $CAUCI_{\alpha}$ metrics. 
%\noindent\textbf{ CelebA:}

\noindent\textbf{Metrics:} We introduced novel metrics, the conjunctive accuracy improvement $CAI_{\alpha}$, a combination of overall accuracy and accuracy gap, and the compound AUC improvement, a combination of AUC and AUC gap, or $CAUCI_{\alpha}$. As our series of experiments on tabular and image data had demonstrated, these novel metrics were generally useful in reflecting wholistic improvements in fairness. For situations where each individual metric was the best, our new metrics reflect this case of Pareto optimality. For more ambiguous cases where one was superior while another was not, such as the accuracy metrics in Table~\ref{table:CelebA_Skincolor_Age}, these examples show the need for a single metric that can help assess best overall performance but also reflect the desired policy objective and ethical imperatives. Table~\ref{tbl:2016_balanced_results_gender} shows a situation where the specific value of $\alpha$ changed the ranking of the methods, indicating the importance of what $\alpha$ is set to. Regarding AUC metrics, showing the effect of methods on the AUC of individual subpopulations appears to be an important consideration for certain scenarios. Unlike accuracy, AUC is not decomposed directly into the AUC on subpopulations, as the AUC represents the capacity of the classifier to choose a specific true positive rate and false positive rate. Consequently, the AUC gap (measuring the disparity in tuning each subpopulation) and minimum AUC (indicating how restricted the worse-off calibration was) were suited to cases where protected factor ground-truth exists in order to calibrate each subpopulation individually. While $CAI_{\alpha}$ and $CAUCI_{\alpha}$ may indeed be useful, guidance for how to set the parameter $\alpha$, or how it relates to legal concepts such as Disparate Impact is deferred to future work.

{\bf Alternative GANs:} 
Pix2pix~\cite{pix2pix2017, pix2pix2019} performs image translation to address domain adaption, but requires pairs of images from the two domains. CycleGAN~\cite{cyclegan2017}, however, allows for style transfer and image translation in the case of unpaired images from two separate domains. The method we propose of latent space manipulation could also be interpreted as working in the unpaired image case, since it uses gradient descent on the loss of a binary classifier that is trained on a cohort of unpaired examples of the two classes (having vs. not having a particular attribute like ethnicity or disease). CycleGAN, in particular, does this by using a pair of coupled GANs, where one is generating images from domain A into domain B and the other going in the other directions. The two GANs are then coupled via their loss functions to achieve cycle consistency. StarGAN~\cite{stargan2018} extends the domain-to-domain adaptation to multi-domain-to-multi-domain adaption by training a single model on multiple domains at once to achieve translation. Like CycleGAN it uses a loss term that expresses cycle-consistency and reconstruction loss as one of its loss terms. The two other loss terms are related to the domain discrimination loss and the traditional GAN image fidelity adversarial loss to make generated images not distinguishable from real images.

Alternatively, the conditional StyleGAN~\cite{oeldorf2019loganv2} could have been used in place of the unconditional StyleGAN we selected. While the conditional StyleGAN has the potential benefit of conditioning the generator on the protected factor, it would limit our understanding of how well the missing factor was supervised, which is why we chose to use the unconditional StyleGAN. There is a possibility for the conditional StyleGAN to produce unrealistic examples; should this be the case, we would have little recourse to correct the generator. However, the unconditional StyleGAN allows for superior control over the protected factors using the style vector; should these results be unrealistic, we can make corrections. The unconditional StyleGAN's generator also has the added benefit of being reusable to control multiple sensitive attributes.

\noindent\textbf{Future Work:} Future work will explore the fact that \AD{} may still produce biased results with regard to a protected factor that was not tested against or has yet to be considered protected. As \IA{} is independent of the downstream task, expanding to other domains, such as segmentation, is also of interest.

Finally, note that our current approach to domain adaptation takes a synthetic sample as a starting point. Our approach could, however, also work by starting with a real sample. One consideration when starting with a real sample is the inversion of this real image from image to latent space, a task that is still under investigation by the research community. The use of StyleGAN2~\cite{Karras2020stylegan2} is a natural option since it enables this inversion, but our preliminary experimentation has shown some limitations in applying the method to non-canonical image types (images more akin to those found in ImageNet rather than retinal or face images). The survey study~\cite{caton2020fairness} explores the lay of the land for such inversion methods and also offers alternative directions for domain adaption, which we intend to explore in future extensions of this work.

\section{Conclusion}

This study proposed TARA, a novel approach to debiasing using joint alteration of data representation and training, aiming to address both sources of bias, conditional dependence and data imbalance. We showed that it outperformed competing methods. We introduced novel fairness metrics addressing some issues in current bias metrics, as a basis for future investigations and discussions between AI scientists, ethicists and policy makers regarding how to best compare and assess debiasing.

\newpage
\section*{Appendix}

\begin{table}[tbh!]
\tiny
    \centering
    \caption{Methods Summary}
    \begin{tabular}{cccc}
    \toprule
    Approach & Goal & Methods & Application \\
    \midrule
        Adversarial Independence & Ensure conditional independence from a protected factor & Adversarial learning & Tabular records and Images \\
        Intelligent Augmentation & Mitigate data imbalance for protected subpopulations & Generative methods + latent space manipulation & Images \\
        TARA & Mitigate data imbalance and ensure conditional independence & Generative methods and Adversarial learning & Images \\
    \bottomrule
    \end{tabular}
    \label{tbl:Methods_Summary}
\end{table}

Additional details on fairness definitions; nomenclature; methods; datasets; preprocessing and implementation; shared code and data; and supplemental discussion items are described below.

\noindent\textbf{Methods Summary:}

A summary of our methods is described in Table~\ref{tbl:Methods_Summary}. 

\noindent\textbf{Nomenclature and Definitions:}
The following includes some of the most commonly used formal definitions of fairness (for more, see also~\cite{mehrabi2019survey} ).

We denote the protected factor(s) as $S$, the classifier's decision for the outcome as $\hat{Y}$, and the true outcome or the underlying true label, depending on context, as $Y$. We focus on the following definitions of fairness~\cite{hardt2016}:

\noindent{\em Demographic Parity:}
Demographic parity states that all subpopulations should have a positive decision (e.g., credit approval) at equal rates. Mathematically, demographic parity states that:
\begin{equation}
    P(\hat{Y} = \hat{y} | S = s) = P(\hat{Y} = \hat{y}) , \forall s, \hat{y}.
\end{equation}
Demographic parity may not be appropriate in situations where a fundamental correlation exists between $Y$ and $S$: consider, for example, a health condition that is predominant in certain age groups, e.g., age-related macular degeneration~\cite{burlina2011automatic}.

\noindent{\em Equality of Odds:}
On the other hand, equality of odds expresses that a predictive model must produce predictions that are conditionally independent of protected factors given the true outcome:
\begin{equation}
P(\hat{Y} = \hat{y} | S = s, Y = y)=P(\hat{Y} = \hat{y} | Y = y), \forall s, y, \hat{y}.
\label{eq:odds}
\end{equation}
Unlike demographic parity, the conditional independence ensures that, when $Y$ has a causal relationship with $S$, the performance of the prediction being correct ($\hat{Y}=Y$) is not adversely affected by the strict condition of independence.

\noindent{\em Equality of Opportunity:}
Equality of opportunity further relaxes the equality constraints in Eq.~\ref{eq:odds} by dictating that a model must produce predictions that are independent of a protected factor, for a specific value $y$ (and not necessarily all) of the true label: 
\begin{equation}\label{eq:opportunity}
   P(\hat{Y} = \hat{y} | Y = y)= P(\hat{Y} = \hat{y} | S = s, Y = y), \forall s, \hat{y}.
\end{equation} 
\noindent{\em Equality of Performance:}
Conditional independence stated above has corollary implications for performance (error rates) of the classifier. Take the example of a binary classification problem; then Eq.~\ref{eq:opportunity}, when stated for $y=1$, ensures an equal {\it true positive} rate exists across all protected factors values $s$. However, unlike equality of odds, it does not necessarily require an equal {\it false negative} rate across all $s$ values. Equality of Odds however does. Demographic parity, equality of odds, and equality of opportunity have served as the foundation for recent advances in AI bias mitigation. For our study we adopt the stricter goals of equality of odds and measure success using commonly adopted metrics as well as novel proposed metrics that are consistent with this goal. 

%\vspace{0.2cm}
Next, we provide extended details on each dataset.

\noindent\textbf{OSMI Mental Health Data Details:}
The OSMI Mental Health in Tech Survey 2014~\cite{KaggleOSMI2014dataset} was released on Kaggle to encourage evaluation of mental health in technology industry and how mental health relates to job related factors.~\cite{rado2019selection} and~\cite{sharma2018predictive} used the OSMI Mental Health Survey 2014 to predict the likelihood a given individual had sought mental health treatment. Approaches evaluated ranged from decision trees to neural networks. Both studies claimed accuracies ranging from 79\% to 98\%. However, evaluation datasets were not standardized across either work. In 2016, OSMI compiled a new mental health survey which included a more extensive questionnaire and more samples. The 2016 OSMI dataset~\cite{KaggleOSMIdataset} included questions asking whether a person has been diagnosed with a mental illness, if so which mental illness, and whether they had sought treatment for a mental illness.~\cite{reddy2018machine} developed a series of off-the-shelf machine learning models that used this dataset to try to predict whether an employee had treatment for mental health related disorders in the past. 

Our work explores the gender and age bias present in the OSMI Mental Health in Tech Survey 2016 dataset (denoted as OSMI) and whether deep learning models can be trained to mitigate bias using \AD{}. For the binary classification task of estimating whether a person sought treatment for a mental illness the prediction network $F$ trained using the set of tabular input features, $X$, listed in Table~\ref{tbl:Mental_Health_Features}. Unlike many of the employer-specific features from the OSMI dataset, these features were selected because they best corresponded to the task of estimating if a person sought mental health treatment. We did, however, ignore features related to personal or family history of mental illness as they were both overly correlated to the likelihood of a person seeking treatment according to~\cite{rado2019selection} and~\cite{sharma2018predictive}.

\begin{table}[tbh!]
\scriptsize
    \begin{center}
\caption{OSMI Features}
\label{tbl:Mental_Health_Features}
        \begin{tabular}{ll}
        \toprule
            Mental Health Feature & \# Classes \\
        \hline
        \midrule
            Age & 53 \\
            Gender & 2 \\
            Benefits and insurance coverage & 2 \\
            Care options & 2 \\
            Anonymous discussions & 2 \\
            Interference with work performance & 2 \\
            Medical leave availability & 2 \\ 
            Perceived negative impact of discussing mental health & 2 \\
        \bottomrule
        \end{tabular}
    \end{center}
\end{table}

\noindent{\bf ITA Note:} 
As the ITA is computed per pixel, care was needed in determining which areas were used to calculate it, as it might have been adversely be affected by light artifacts and lighting in general.

\noindent\textbf{EyePACs Data Details:}
The EyePACs dataset contained 88,692 images for 44,346 participants, with two images, the left and right fundus, for each participant. We resized these images to 256x256 pixels after being cropped to the outline of the fundus, and the labels binarized such that 0 and 1 were "not referable" and 2, 3, and 4 were "referable" for the disease. To compute the mask for ITA on each image, we ran a one-class SVM, with a RBF kernel and a upper bound of 80\% for the training errors, on the luminance dimension, with each non-background pixel as a data point, to mask any anomalous areas such as light artifacts along with the background. The ITA was then computed per pixel and averaged over the non-masked area.

\noindent\textbf{CelebA Data Details:}
These images were 218x178 pixels, and were preprocessed by taking a 128x128 crop with the center at (121, 89). Older dark skinned females were the smallest subpopulation consisting of 1,380 images, where an ITA less than 28 denoted dark skinned, whereas younger light skinned females were the largest at 93,477 images.

We computed the ITA in a similar manner to \cite{merler2019diversity}, where we used a skin segmentation step to filter out invalid pixels, and used a landmark detector to segment out the chin, cheeks and forehead areas. We diverged slightly in that we used Gaussian blur on the image of ITA values (with a kernel size of 11) and we chose the median ITA value per region that we then averaged over to get the final value.

\begin{table}[htb!]
\tiny
    \begin{center}
\caption{OSMI Demographic parity results. Performance metrics for debiasing methods on OSMI predicting $Y$= sought mental health treatment, trained on partitioning with respect to $S$ = Gender, and evaluated on a test set balanced across treatment status and gender (M/F). Methods include: FC network (Cat), FC with embeddings network (EMB), noise debias (Noise), adversarial debias with demographic parity (ADDP), adversarial debias with equality of odds (AD), and freeze training (Freeze).}\label{tbl:2016_demo_parity_results_gender}
        \begin{tabular}{l||c|c|c|c|c|c|c}
        \toprule

            Metrics & Baseline (EMB) & Cat & EMB+Noise &  Freeze EMB+ADDP &  Cat+ADDP & EMB+ADDP & EMB+AD \\

        \hline
        \midrule
            $acc$ & 63.22 (5.07) & 50.57 (5.25) & 67.82 (4.91) & 76.15 (4.48) & 56.03 (5.21) & 77.59 (4.38) & \textbf{82.18 (4.02)} \\
            \hline
            
            $acc_{gap}$ & 20.69 (0.87) & \textbf{2.29 (0.01)} & 24.14 (1.42) & 6.32 (0.58) & 2.87 (0.05) & 3.45 (0.34) & 4.59 (0.58) \\
            \hline
            
            $acc_{min}$ (subpop.) & 52.87 (F) & 49.43 (M) & 55.75 (F) & 72.99 (M) & 54.60 (M) & 75.86 (M) & \textbf{79.89 (M)} \\
            \hline
            
            $CAI_{0.5}$ & - & 2.88 & 0.58 & 13.65 & 5.32 & 15.81 & \textbf{17.53} \\
            \hline
            
            $CAI_{0.75}$ & - & 10.64 & -1.44 & 14.01 & 11.57 & 16.52 & \textbf{16.82} \\
            \hline
            \midrule
            
            $AUC$ & 0.7213 (0.0471) & 0.5807 (0.0518) & 0.8030 (0.0418) & 0.8407 (0.0384) & 0.5905 (0.0517) & \textbf{0.8633 (0.0361)} & 0.8592 (0.0365) \\
            \hline
            
            $AUC_{gap}$ & 0.0606 (0.0087) & 0.0854 (0.0026) & 0.0610 (0.0102) & 0.1030 (0.0146) & 0.0663 (0.0019) & \textbf{0.0537 (0.0085)} & 0.0805 (0.0129) \\
            \hline
            
            $AUC_{min}$ (subpop.) & 0.8171 (M) & 0.5580 (F) & \textbf{0.8430 (M)} & 0.7894 (M) & 0.5615 (M) & 0.8380 (M) & 0.8236 (M) \\
            \hline
           
            $CAUCI_{0.5}$ & - & -0.0827 & 0.0407 & 0.0385 & -0.0683 & \textbf{0.0745} & 0.0590 \\
            \hline
            
            $CAUCI_{0.75}$ & - & -0.0538 & 0.0201 & -0.0020 & -0.0370 & \textbf{0.0407} & 0.0196 \\
        \bottomrule
        \end{tabular}
    \end{center}
    %\vspace{-0.6cm}
\end{table}

\begin{table}[tbh!]
\tiny
    \begin{center}
\caption{OSMI Demographic parity results. Performance metrics for debiasing methods on OSMI predicting $Y$= sought mental health treatment, trained on partitioning with respect to $S$ = Age, and evaluated on a test set balanced across treatment status and age (O/Y for Older/Younger).}\label{tbl:2016_demo_parity_results_age}
        \begin{tabular}{l||c|c|c|c|c|c|c}
        \toprule

            Metrics & Baseline (EMB) & Cat & EMB+Noise &  Freeze EMB+ADDP &  Cat+ADDP & EMB+ADDP & EMB+AD \\

        \hline
        \midrule
            $acc$ & 68.36 (4.29) & 63.05 (4.45) & 75.00 (3.99) & 73.89 (4.05) & 57.08 (4.56) & 79.42 (3.73) & \textbf{80.75 (3.63)} \\
            \hline
            
            $acc_{gap}$ & 21.68 (1.15) & 19.03 (0.68) & 18.14 (1.41) & 16.81 (1.22) & 12.39 (0.24) & 3.99 (0.38) & \textbf{1.33 (0.14)} \\
            \hline
            
            $acc_{min}$ (subpop.) & 57.52 (O) & 53.54 (O) & 65.93 (Y) & 65.49 (O) & 50.88 (Y) & 77.43 (O) & \textbf{80.09 (O)} \\
            \hline
            
            $CAI_{0.5}$ & - & -1.33 & 5.09 & 5.20 & -1.00 & 14.38 & \textbf{16.37} \\
            \hline
            
            $CAI_{0.75}$ & - & 0.66 & 4.32 & 5.04 & 4.15 & 16.03 & \textbf{18.36} \\
            \hline
            \midrule
            
            $AUC$ & 0.7658 (0.0444) & 0.7249 (0.0412) & 0.8211 (0.0353) & 0.8458 (0.0333) & 0.6171 (0.0448) & \textbf{0.8928 (0.0285)} & 0.8787 (0.0301) \\
            \hline

            $AUC_{gap}$ & 0.0014 (0.0002) & 0.0360 (0.0030) & \textbf{0.0003 (0.0001)} & 0.0090 (0.0014) & 0.0636 (0.0024) & 0.0310 (0.0054) & 0.0139 (0.0021) \\
            \hline

            $AUC_{min}$ (subpop.) & 0.8648 (O) & 0.7567 (Y) & 0.8780 (Y) & 0.8754 (Y) & 0.6059 (Y) & \textbf{0.8826 (Y)} & 0.8761 (Y) \\
            \hline

            $CAUCI_{0.5}$ & - & -0.03775 & 0.0282 & 0.0362 & -0.1055 & 0.0487 & \textbf{0.0502} \\
            \hline
            
            $CAUCI_{0.75}$ & - & -0.0362 & 0.0147 & 0.0143 & -0.0838 & 0.0096 & \textbf{0.0189} \\
        \bottomrule
        \end{tabular}
    \end{center}
    %\vspace{-0.6cm}
\end{table}

%\vspace{0.2cm}

In addition to what was reported in the main body of the paper, we also performed more experiments on OSMI, which provided additional insights in the workings of \AD{}. These are detailed next.

\noindent {\bf Extended OSMI Experiments and Discussion:}

We performed an ablation study on the network architecture depicted in Figure~\ref{fig:adv_debias_network_diagram}, which we evaluated on the OSMI dataset to determine the impact of \AD{} (without \IA{}) on fairness. We experimented with replacing the embedding layers (methods containing {\it Cat}) and removing the adversarial module altogether (without {\it AD} or {\it ADDP}). The {\it Cat} methods concatenated each of the input features into a single fully connected layer. Our proposed \AD{} debiasing approach ({\it EMB+AD}) included a prediction network with an embedding for each of the input features concatenated and forwarded to a fully connected layer, while the adversarial debiasing network was a single fully connected layer ({\it AD}). Moreover, we also evaluated an alternative adversarial module constructed based on demographic parity ({\it ADDP}) instead of equality of odds, meaning the adversarial module only received the prediction module's logits as an input and did not use the prediction task's target label. 

First, we examined the gender debiasing results shown in Table \ref{tbl:2016_demo_parity_results_gender} reflecting a 12.65\% increase in overall accuracy when using and embedding-based prediction network ({\it EMB}) compared with the concatenation network without adversarial debiasing ({\it Cat}) performance. However, the {\it Cat} network had the smallest accuracy gap among all methods evaluated, but at the cost of near random accuracy. All the {\it Cat}-based methods suffered from poor accuracy which is likely attributed to generalizing poorly to the unseen class (females seeking treatment) due to the limited network structure (a single fully connected layer), unlike the {\it EMB}-based methods which also had the contribution of embedding layers. Together these deficiencies in the {\it Cat}-based methods were visible in both the $CAI_{\alpha}$ and $CAUCI_{\alpha}$ being the lowest. These same experiments were also conducted for the case debiasing the age protected factor (shown in Table \ref{tbl:2016_demo_parity_results_age}), which also showed the importance of using embeddings for OSMI. Overall, embedded features ({\it EMB} methods) maximized overall accuracy, AUC, and minimum accuracy of the subpopulations (in this case female) that was under represented. 

Next, the impact of using \AD{} based on demographic parity ({\it ADDP}) instead of equality of odds ({\it AD}) was examined. The results suggest that {\it ADDP} was almost as effective as {\it AD} in terms of accuracy-based metrics, as best represented in the $CAI_{\alpha}$ for $\alpha = 0.5$ and $0.75$. However, for debiasing gender the {\it ADDP} was superior to the {\it AD} method in terms of the $CAUCI_{\alpha}$. Similarly, for the case of debiasing age the {\it EMB+AD} and {\it EMB+ADDP} were competitive with one another with respect to the AUC-based metrics. To better understand which method performed better overall in terms of AUC, we examined the compound AUC improvement ($CAUCI_{\alpha}$) which showed {\it EMB+AD} was marginally better for $\alpha=0.5$ and substantially better for $\alpha=0.75$, which weighed the AUC gap more heavily. The results for gender and age debiasing suggested that using adversarial learning based on equality of odds ({\it AD}) to reduce the prediction network's bias towards a protected class could reduce bias and encourage it to identify more optimal network weights.

Last, we evaluated the impact of an alternative to training the adversarial debiasing module, where the adversarial network ({\it ADDP}) was frozen when training the prediction network and then the prediction network was frozen when training the adversarial network (denoted as {\it Freeze EMB+ADDP}). Alternating freeze training allowed each network to be optimized individually in order to potentially improve individual performance without affecting the other. However, the results in Tables \ref{tbl:2016_demo_parity_results_gender} and \ref{tbl:2016_demo_parity_results_age} indicate that freeze training performed worse than non-freeze training in $CAUCI_{\alpha}$ and $CAI_{\alpha}$ for debiasing gender and age. As a result, optimizing both the prediction and adversarial networks without freezing parameters was shown to be preferable.

\noindent {\bf OSMI Experiments Implementation Details:}
OSMI was partitioned by randomly shuffling and spliting into 70\%, 10\%, and 20\% partitions corresponding to train, validation, and test sets, respectively. As mentioned in Section~\ref{sec:Experiments} (Domain Generalization), categories that had the fewest members (Gender debiasing = female seeking treatment, Age debiasing = older seeking treatment) were relegated to the test partition only. The test split was constructed in a manner where each member of the protected class had equal representation. Each of the models being evaluated were trained on the respective dataset for up to 100 epochs, with early stopping triggering when the validation loss had not decreased for 10 epochs. After some experimentation, we found that the adversarial loss balancing term $\beta$ resulted in the best performance when set to 1. The network weights corresponding to the smallest validation loss were retained for evaluation on the test set. We used the Adam optimizer with a learning rate scheduler set to reduce the learning rate by a factor of 0.1 for every 10 epochs the training loss plateaued. All adversarial modules were pre-trained for 100 epochs with the prediction module frozen. The pre-training procedure was designed to help reduce the likelihood of poor initialization of network weights for the debiasing module.

\noindent\textbf{Image-based Experiments Implementation details:}
For image experiments, we used a ResNet50 classifier pretrained on ImageNet with the final linear layer replaced with a randomly initialized layer with an output dimension of 2. The adversarial network consisted of four fully connected layers of width 512 with LeakyRelu activations and $\alpha = 0.01$. We used stopping criterion on the lowest validation loss, with a patience of 5 epochs for all methods. Outside of dataset specific preprocessing, we used Imagenet normalization on the input, and resized to 224x224 using a bicubic interpolation. Additionally, we used SGD with a learning rate of $\gamma=0.001$ and Nesterov moment of 0.9, and AdamW with a learning rate of 0.005 for the adversary.

\noindent\textbf{EyePACs Experiments:}
 Our training dataset was made up of 10,346 referable lighter skin images ($ITA=0, DR=1$), 5,173 non-referable lighter skin images ($ITA=0, DR=0$) and 5,173 non-referable darker skinned images ($ITA=1, DR=0$). One finding of note was that the lighter skin subpopulation transitioned from performing better than the darker skin subpopulation to performing worse when using \AD{} methods without filtering.

\noindent\textbf{CelebA Experiments:}
We used a total training dataset size of 48,000 for each partitioning, where we again kept each training dataset balanced across our target factor, age, and our protected factors. For gender, there were 24,000 older male images, 12,000 younger male and 12,000 younger female images. For ITA, there were 24,000 older dark skinned images, 12,000 younger lighter skinned images, and 12,000 younger darker skinned images.

\noindent\textbf{Metrics:}
Note, in some cases the minimum accuracy performance flipped between subpopulations for the baseline versus the methods with \AD{}. Given this occurrence, despite the minimum accuracy avoiding selection of a coefficient $\alpha$, compared to $CAI_{alpha}$, it nevertheless still requires scrutiny from ethicists and policy makers to ensure the metric does not give false understanding of which subpopulation is underrepresented.

\end{document}